\documentclass[10pt,journal,compsoc]{IEEEtran}

\ifCLASSOPTIONcompsoc
  \usepackage[nocompress]{cite}
\else
  \usepackage{cite}
\fi

\usepackage{amsmath,amsfonts}
\usepackage{pifont}
\usepackage{array}
\usepackage[caption=false,font=normalsize,labelfont=sf,textfont=sf]{subfig}
\usepackage{textcomp}
\usepackage{stfloats}
\usepackage{url}
\usepackage{verbatim}
\usepackage{graphicx}
\usepackage{cite}
\usepackage{epsfig}
\usepackage{amssymb}
\usepackage{booktabs}
\usepackage{enumitem}
\usepackage{multirow}
\usepackage{lipsum}
\usepackage{xcolor}
\usepackage[export]{adjustbox}
\definecolor{codegreen}{rgb}{0.0,0.6,0.0}

\usepackage[ruled,vlined,linesnumbered]{algorithm2e}
\makeatletter
\newcommand{\algorithmfootnote}[2][\footnotesize]{%
  \let\old@algocf@finish\@algocf@finish
  \def\@algocf@finish{\old@algocf@finish
    \leavevmode\rlap{\begin{minipage}{\linewidth}
    #1#2
    \end{minipage}}%
  }%
}
\makeatother
\makeatletter
\DeclareRobustCommand\onedot{\futurelet\@let@token\@onedot}
\def\@onedot{\ifx\@let@token.\else.\null\fi\xspace}

\def\eg{\emph{e.g}\onedot}

\def\etc{\emph{etc}\onedot}

\makeatother
\usepackage[pagebackref,breaklinks,colorlinks]{hyperref}

\usepackage[capitalize]{cleveref}
\crefname{section}{Sec.}{Secs.}
\Crefname{section}{Section}{Sections}
\Crefname{table}{Table}{Tables}
\crefname{table}{Tab.}{Tabs.}

\newcommand{\lzl}[1]{\textcolor{black}{#1}}

\begin{document}

\title{SparseTrack: Multi-Object Tracking by Performing Scene Decomposition based on Pseudo-Depth}

\author{Zelin Liu,
        Xinggang Wang,
        Cheng Wang,
        Wenyu Liu
        }

\author{Zelin Liu,
        Xinggang Wang,
        Cheng Wang,
        Wenyu Liu,
        Xiang Bai
\IEEEcompsocitemizethanks{
\IEEEcompsocthanksitem Z. Liu, X. Wang, C. Wang, W. Liu and X. Bai are with Huazhong University of Science and Technology, China.\protect\\
\IEEEcompsocthanksitem Corresponding author: Xinggang Wang. Email: \url{xgwang@hust.edu.cn}.\protect\\
}
}

\markboth{Journal of \LaTeX\ Class Files,~Vol.~14, No.~8, August~2021}%
{Shell \MakeLowercase{\textit{et al.}}: A Sample Article Using IEEEtran.cls for IEEE Journals}

\IEEEpubid{0000--0000/00\$00.00~\copyright~2021 IEEE}

\IEEEtitleabstractindextext{%
\begin{abstract}
    Exploring robust and efficient association methods has always been an important issue in multiple-object tracking (MOT). Although existing tracking methods have achieved impressive performance, congestion and frequent occlusions still pose challenging problems in multi-object tracking. We reveal that performing sparse decomposition on dense scenes is a crucial step to enhance the performance of associating occluded targets. To this end, we propose a pseudo-depth estimation method for obtaining the relative depth of targets from 2D images.  Secondly, we design a depth cascading matching (DCM) algorithm, which can use the obtained depth information to convert a dense target set into multiple sparse target subsets and perform data association on these sparse target subsets in order from near to far. By integrating the pseudo-depth method and the DCM strategy into the data association process, we propose a new tracker, called SparseTrack. SparseTrack provides a new perspective for solving the challenging crowded scene MOT problem. Only using IoU matching, SparseTrack achieves comparable performance with the state-of-the-art (SOTA) methods on the MOT17 and MOT20 benchmarks. Code and models are publicly available at \url{https://github.com/hustvl/SparseTrack}.
\end{abstract}

}

\maketitle
\IEEEdisplaynontitleabstractindextext

\section{Introduction}\label{intro}
\IEEEPARstart{M}{ulti}-object tracking (MOT)~\cite{vandenhende2021multi} has vast applications in fields such as autonomous driving, surveillance, and intelligent transportation. It aims to consistently identify the same object in different video frames as the same identity in the form of bounding boxes. Although previous trackers have achieved high performance on multiple tracking datasets \cite{mot15, mot16, mot20}, dense crowds and frequent occlusions still make multi-object tracking tasks challenging.

\begin{figure}[!t]
\centering
\includegraphics[width=1.0\linewidth]{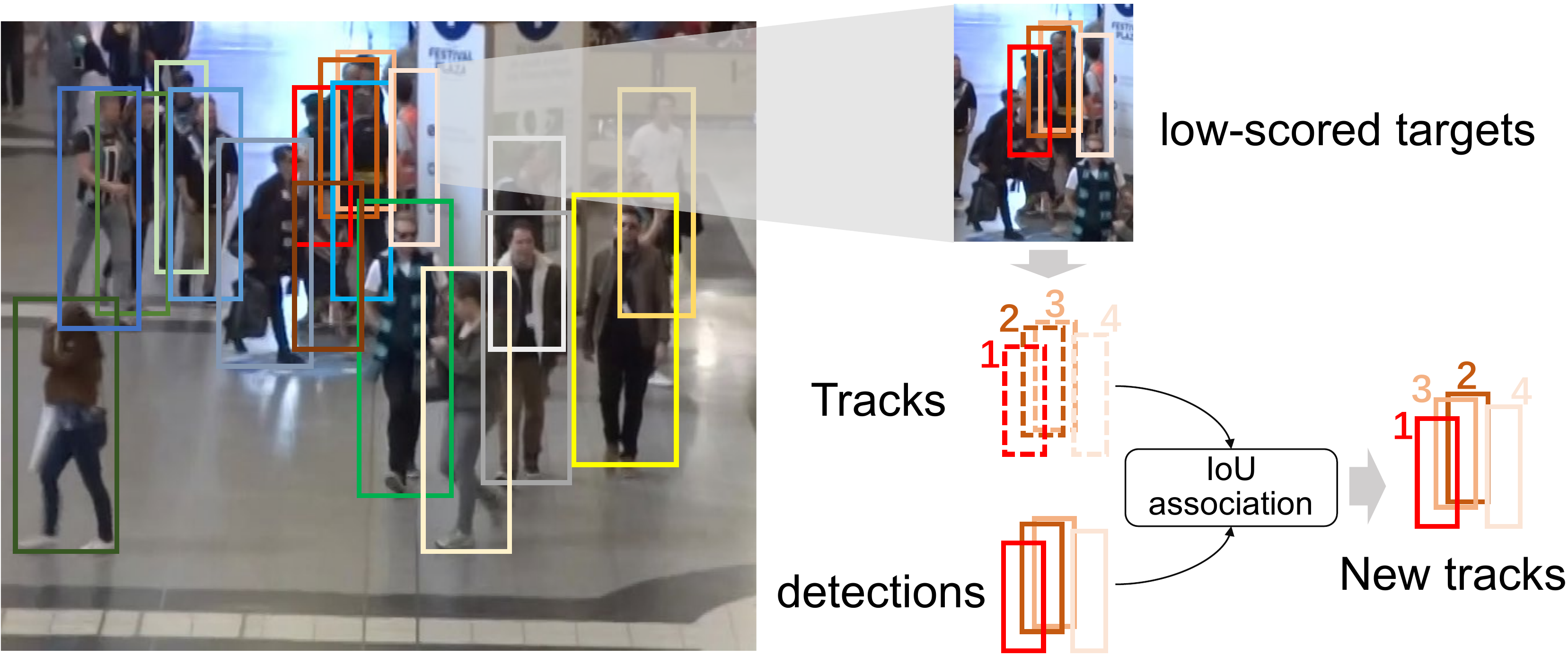}
\caption{An illustration of associating low-score detections in crowded scenes. In ByteTrack \cite{bytetrack},  a set of low-score detections are matched with the track set at the same time using the IoU metric, which is easy to make mistakes due to the high location similarity between the low-score detections.  This paper attempts to solve this problem from the low-score detection decomposition perspective rather than matching them at the same time.}
\label{low-dets}
\end{figure}

Current mainstream tracking methods follow the paradigm of tracking-by-detection (TBD)~\cite{Bewley2016_sort} and perform frame-by-frame data association. In order to solve the obstacle of occlusion association in dense scenes, some simple methods, such as ByteTrack~\cite{bytetrack}, have achieved effective tracking of occluded targets in dense scenes by separately associating low-score detections. Although ByteTrack demonstrates proficiency in processing low-score detections separately, its accuracy in location association is prone to deterioration in scenes characterized by a high volume of low-score occlusions or frequent overcrowding, as shown in \cref{low-dets}. Other methods~\cite{transcenter,transtrack,transmot,p3aformer,motr,motrv2} ensure tracking performance of occluded instances by using powerful temporal modeling and trajectory query mechanisms. However, these methods are typically associated with high computational costs, particularly in scenes populated by a multitude of objects and frequent occlusions.

\begin{figure}[!t]
\centering
\includegraphics[width=1.0\linewidth,]{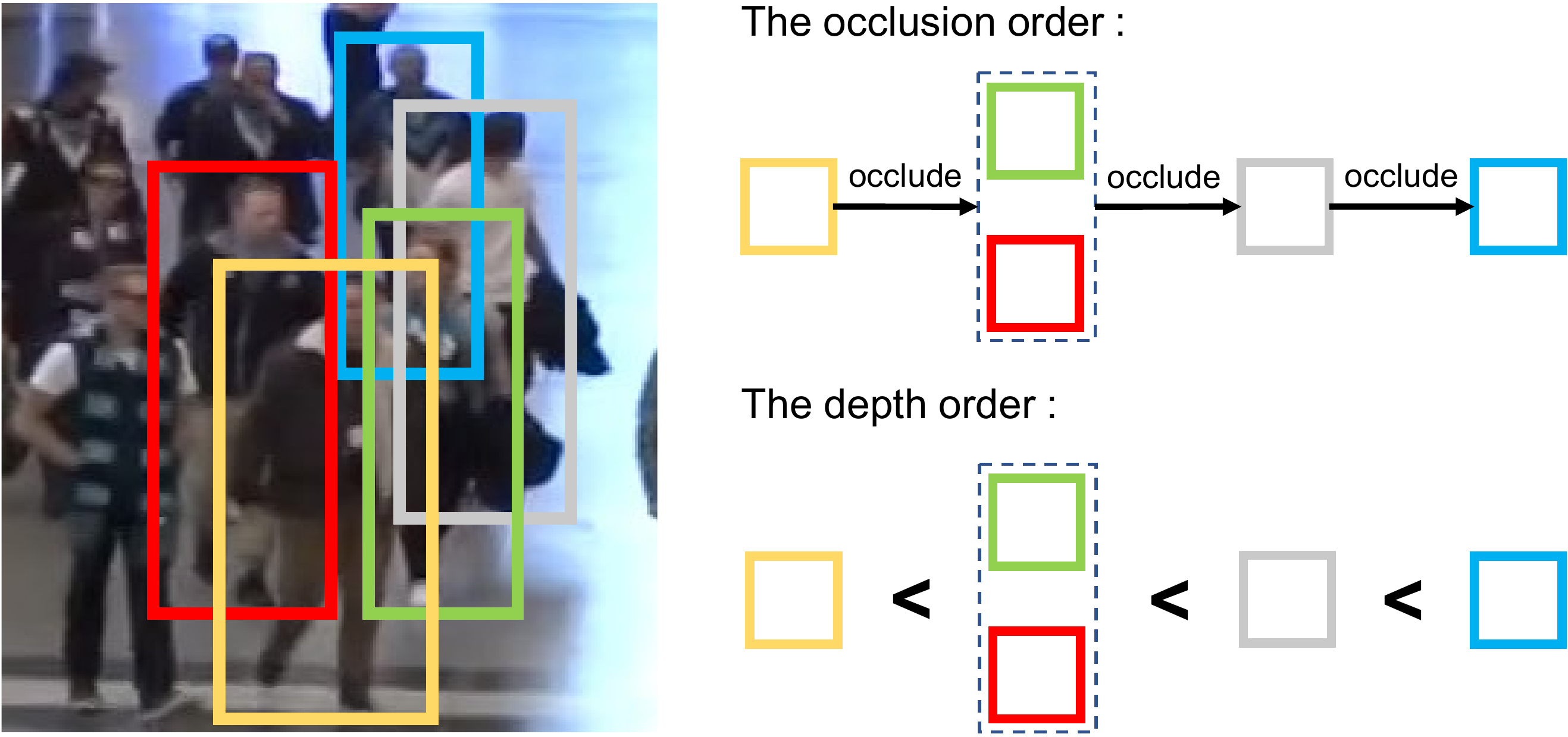}
\caption{An illustration of the occlusions at the local regions. As the depth value increases, the ranking of occluded targets gradually shifts toward the background.}
\label{order}
\end{figure}

In this work, we prove that the target set decomposition based on depth information is an effective approach for dealing with dense occlusions in data association. As illustrated in \cref{order}, we show an occlusion order for the objects in a local region and the occlusion order is consistent with the depth order, from near to far. Thus, a set of dense occlusions can be divided into several non-overlapping subsets by utilizing a segmentation strategy based on depth information. Between adjacent target subsets, their $x$-$y$ locations could be similar, but their $x$-$y$-$depth$ locations can be much easier to distinguish. The tracker performs data association separately for sparse subsets at different depth levels. Compared to directly associating the entire occluded object set at the same time, the sparse decomposition would be more effective for alleviating the collision probability of trajectories with similar positions but different depths during data association. Specifically, we prioritize the association of target subsets with smaller depths since the occlusion order is highly correlated with the depth order. As a result, the targets in each subset can be handled in a fine-grained manner and be less affected by the targets in other depth levels.
\IEEEpubidadjcol

\begin{figure}[!t]
\centering
\includegraphics[width=1.0\linewidth]{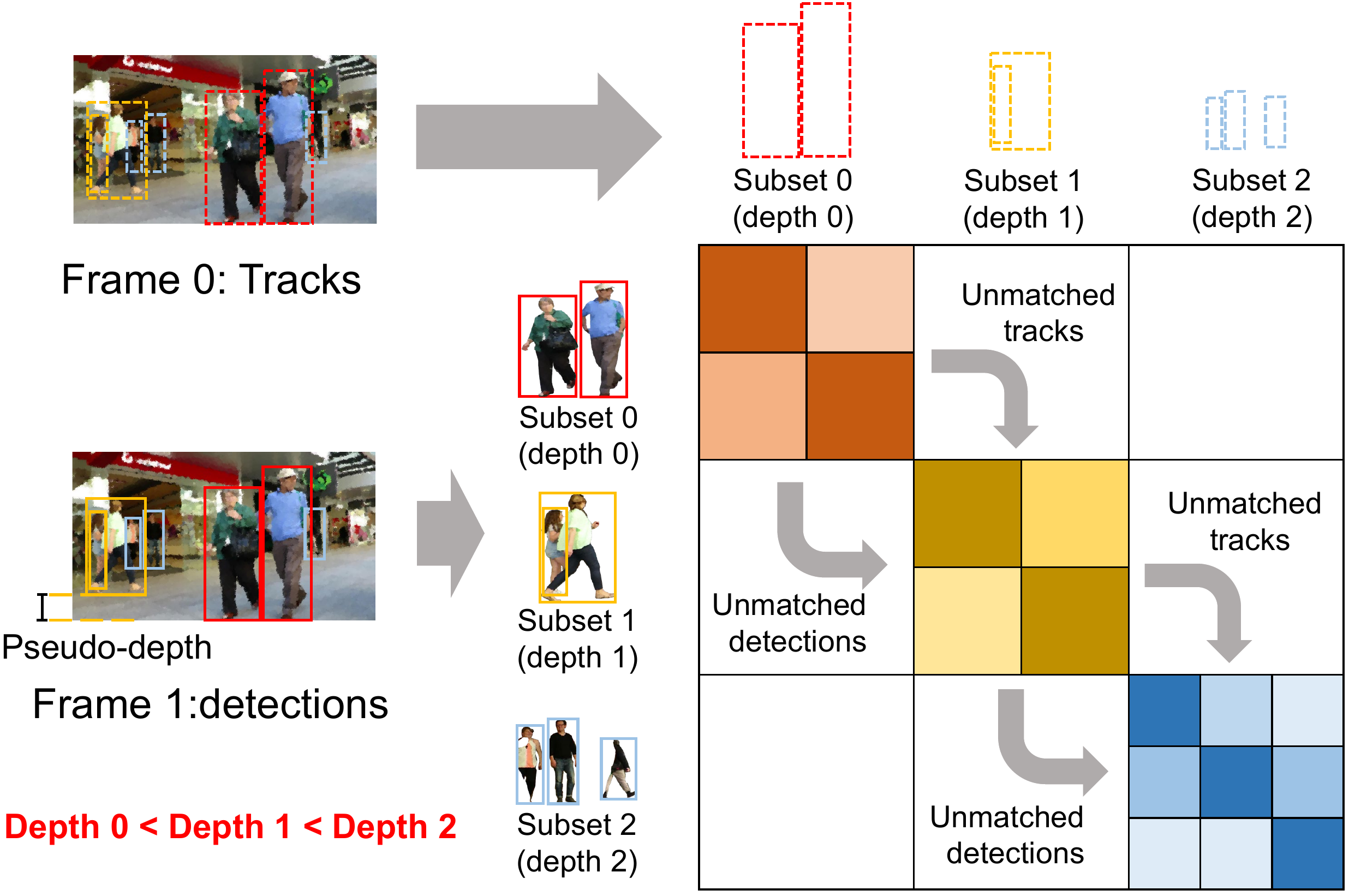}
\caption{An simple example of the hierarchical association from DCM, where the white square regions indicate that no data association is performed between the corresponding detection and trajectory subsets along both horizontal and vertical directions. The darkness of the square regions reflects the similarity between trajectories and detections, with darker colors indicating higher similarity. DCM decomposes the target set into multiple subsets according to depth order from near to far and performs data association on the detection and trajectory subsets at the same depth level. For the unmatched trajectories and detections from each depth level, the tracker will process them at the next depth level.}
\label{sparse}
\end{figure}

To realize the above designs, we propose a method for obtaining the relative depth of targets from 2D images: the pseudo-depth method, which is based on two scene priors: 1) the camera that captures objects is higher than the ground, 2) all objects in the scene are on flat ground. In other words, there are no obvious undulations on the ground. In this case, we can project the relative depth of targets from 3D space onto the 2D image plane and obtain target pseudo-depth values, which is the distance from the bottom of the target bounding box to the bottom edge of the image. Notably, the pseudo-depth value is used as a reference to measure the relative depth relationship between targets using the ground as a reference system, rather than the ground truth depth of the target in 3D space. Furthermore, we design a depth cascade matching (DCM) algorithm to execute hierarchical association based on the target pseudo-depth information. Specifically, we divide trajectories and detections into multiple target subsets according to the distribution of pseudo-depth value. The DCM algorithm performs IoU~\cite{girshick14CVPR} association on these sparse target subsets in the order of pseudo-depth value from nearest to farthest. By integrating the pseudo-depth method and DCM into the data association process, we propose a novel tracker called SparseTrack. The essence of SparseTrack lies in associating occlusions hierarchically based on depth via the DCM, as shown in~\cref{sparse}.

SparseTrack achieves impressive performance on multiple tracking datasets, which proves the effectiveness of target set decomposition based on pseudo-depth. We take ByteTrack as a baseline. On the MOT17~\cite{mot16} test set, SparseTrack achieved 65.1 HOTA~\cite{hota}, 81.0 MOTA, and 80.1 IDF1, which are gains of \textbf{+2.0} HOTA, \textbf{+0.7} MOTA, and \textbf{+2.8} IDF1 compared to the baseline. On the MOT20~\cite{mot20} dataset, SparseTrack achieved 63.4 HOTA, 78.2 MOTA, and 77.3 IDF1, which are gains of \textbf{+2.1} HOTA, \textbf{+0.4} MOTA, and \textbf{+2.1} IDF1 compared to the baseline. Furthermore, we evaluate SparseTrack on DanceTrack~\cite{dancetrack} benchmark and obtain a gain of \textbf{+7.8} HOTA, \textbf{+1.7} MOTA, \textbf{+4.4} IDF1 compared to the baseline. It is worth noting that our proposed DCM algorithm is plug-and-play and can be integrated into different trackers, resulting in consistent performance improvements. The specific details are discussed in the experimental section.

Our contributions are summarized as follows:

\begin{itemize}
\item{We propose a method for obtaining the relative depth of targets from 2D images: the pseudo-depth method, which is based on two prior in the scene, and can effectively obtain the pseudo-depth value of the target to compare the relative depth relationships between different objects.}

\item{Based on the depth information provided by the pseudo-depth method, we design an effective depth cascade matching approach for associating occlusions in dense scenes. It can decompose dense target sets into multiple sparse target subsets to achieve scene decomposition.}

\item{Based on the aforementioned design, we propose a new IoU-only tracker named as SparseTrack, which significantly outperforms the previous IoU-only trackers and achieves comparable results with recent state-of-the-art MOT methods on a wide range of MOT benchmarks.}
\end{itemize}

\section{Related work}\label{related}
\subsection{2D Multi-Object Tracking}
\lzl{The current mainstream work on 2D multi-object tracking focuses on ensuring robust temporal associations, which can lead to the development of various tracking methods. Based on different temporal modeling methods, current mainstream tracking approaches can be roughly divided into four categories: temporal associations via positional information, temporal associations via appearance features, temporal associations via graph optimization, and temporal associations via attention mechanisms. Early trackers \cite{Bewley2016_sort, ioutracker}, based on deep learning perform temporal associations between consecutive frames via positional information and employ the Kalman filter \cite{kf} based on constant velocity motion priors to model the inter-frame motion of the targets, which can ensure the stability of the trajectories. To achieve the long-term association, DeepSORT \cite{deepsort} and MOTDT \cite{motdt} enhance SORT \cite{Bewley2016_sort} by adding appearance cues to compensate for the defect of positional information in associating long-term trajectories. Although appearance cues generated by re-identification (ReID) models \cite{fastreid, torchreid, zhou2019osnet, zhou2021osnet} are effective for associating long-term objects, challenges are posed by target occlusions and motion blurs to the reliability of these cues. Some methods \cite{unitrack,qdtrack,UTM,semi,online_mot} use the ReID models enhanced by self-supervised pre-training \cite{moco,mocov2,simclr,simclrv2,MAE} to alleviate the poor appearance of occluded targets, but it comes with a higher computational cost. Other trackers, such as \cite{jde,fairmot,CStrack}, improve computational efficiency by jointly optimizing the detection task and the re-identification task within a single framework, which somewhat limits the upper bound of detection performance and appearance quality. Partial works \cite{motr,motrv2,transtrack,transcenter,trackformer} are inspired by Transformer \cite{attention,vit,detr,deformable-detr} and attempt to implement trajectory temporal propagation from the perspective of attention mechanism and achieve impressive performance in complex scenes \cite{dancetrack,bdd100k}, While the computational overhead generated by self-attention and feed-forward networks is still non-negligible. Although the methods \cite{dt,centertrack,siammot, Tracktor,ctracker} offer computationally efficient trajectory temporal propagation based on convolutional neural network (CNN) \cite{cnn}, they struggle with the association problem for crowded targets. Recent works \cite{transmot,sushi, graph-mot, Learnable_graph_matching, graph-mot1, GSDT}, formulate multiple object tracking as a graph optimization problem, using graph neural networks or graph matching to achieve robust data association. However, the computational overhead based on graphs tends to increase square fold when the number of targets in the scene increases.}

\subsection{Methods for Handling Object Occlusion}
The performance of a tracker is to some extent dependent on the ability to handle occlusions. Recent works have attempted to tackle occlusion from different perspectives. For example, MotionTrack \cite{motiontrack} learns the motion pattern of trajectories and combines it with historical information to effectively model tracks of occlusions. MOTR \cite{motr}, an end-to-end MOT framework, processes new appearing targets and tracked targets separately using detection queries and trajectory queries, respectively, with a multi-frame training manner. Due to the relative independence among queries and sufficient temporal training, MOTR can perform well in temporal modeling of occlusions across time steps. \lzl{DP-MOT \cite{dp_mot} proposes a subject-ordered depth estimation (SODE) method to automatically sort the depth positions of detected objects in 2D scenes in an unsupervised manner. By constructing a pseudo-3D Kalman filter, DP-MOT achieves robust association for occluded targets.} BoT-SORT \cite{BoT-SORT} combines camera motion compensation (CMC) and IoU-ReID fusion, which can integrate motion and appearance clues of the object for achieving accurate tracking of occlusions. \lzl{OUTrack \cite{online_mot} uses an unsupervised re-identification module and occlusion-aware module to predict the locations where target occlusions occur, in order to compensate for missed detections. ApLift \cite{aplift} introduces an advanced approximate solver for tackling the disjoint path problem, specifically engineered to manage extended and congested trajectory sequences, with performance comparable to mainstream tracking methods. \cite{9857481} segment the trajectories of pairwise occluded targets and recalculates trajectory similarities to effectively associate occluded targets.} SparseTrack provides a completely different solution for associating occlusions by performing the target set decomposition to align the location interval of occlusions.

\subsection{Targets Set Decomposition}
Actually, many trackers that follow the tracking-by-detection paradigm perform some degree of the target set decomposition. For example, DeepSORT \cite{deepsort} achieves step-by-step association of the detection set through multi-stage cascaded matching. FairMOT \cite{fairmot} processes the detection set separately based on appearance and position clues, which can also be seen as a division of the detection set via tracking clues. \lzl{LMGP \cite{LMGP} effectively eliminates trajectory errors in single-camera trackers by introducing a pre-clustering method driven by 3D geometric projections, which groups detected objects for association.} ByteTrack \cite{bytetrack} divides the detection set into high-score and low-score detections based on confidence scores, and uses the correlation between scores and occlusions to separately process low-score occluded targets. However, in dense crowds, congestion leads to occlusions, which results in low-score detections. Although ByteTrack decouples low-score occlusions from the scene, low-score targets are still crowded. In this case, IoU-based data association is prone to matching errors that limit the tracking ability to handle occlusions. To this end, SparseTrack divides the detection set and the trajectory set based on the pseudo-depth level to ensure that targets in each association step are no longer crowded. 

\subsection{The Applications of Depth Information in MOT}
Depth information is commonly used in applications related to 3D scenes, such as monocular or stereo depth estimation, 3D detection and tracking, and recently popular neural radiance fields \cite{nerf}. The standard depth provides abundant information on the appearance and position of objects, leading to different 3D tracking methods. For instance, TrackRCNN \cite{trackrcnn} uses appearance and motion features for tracking and enhances accuracy and stability by alternately tracking objects in 2D images and 3D point clouds. FANTrack \cite{FANTrack} employs a feature-based association network (FAN) for object tracking, where features come from a 3D convolutional neural network (CNN) \cite{cnn} for point cloud data and a 2D CNN for camera images. CenterPoint \cite{centerpoint} uses a keypoint detector to detect the center of an object and regress to other attributes, including 3D size, 3D direction, and speed. Then, it performs 3D object tracking via simple greedy nearest point matching. In fact, a 2D image can be considered as the projection of a 3D scene under perspective transformation. According to the camera model, we can easily infer the variable relationship between a 2D image and a 3D scene. SparseTrack leverages this variable relationship to obtain the pseudo-depth values of targets in 2D image, which partially replace the role of standard depth in describing the position relationships among objects. To the best of our knowledge, SparseTrack is the first method that utilizes depth information to \lzl{decompose targets set.}

\section{method}\label{meth}
\begin{figure*}[!t]
\centering
\includegraphics[width=0.97\linewidth, height=9.5cm]{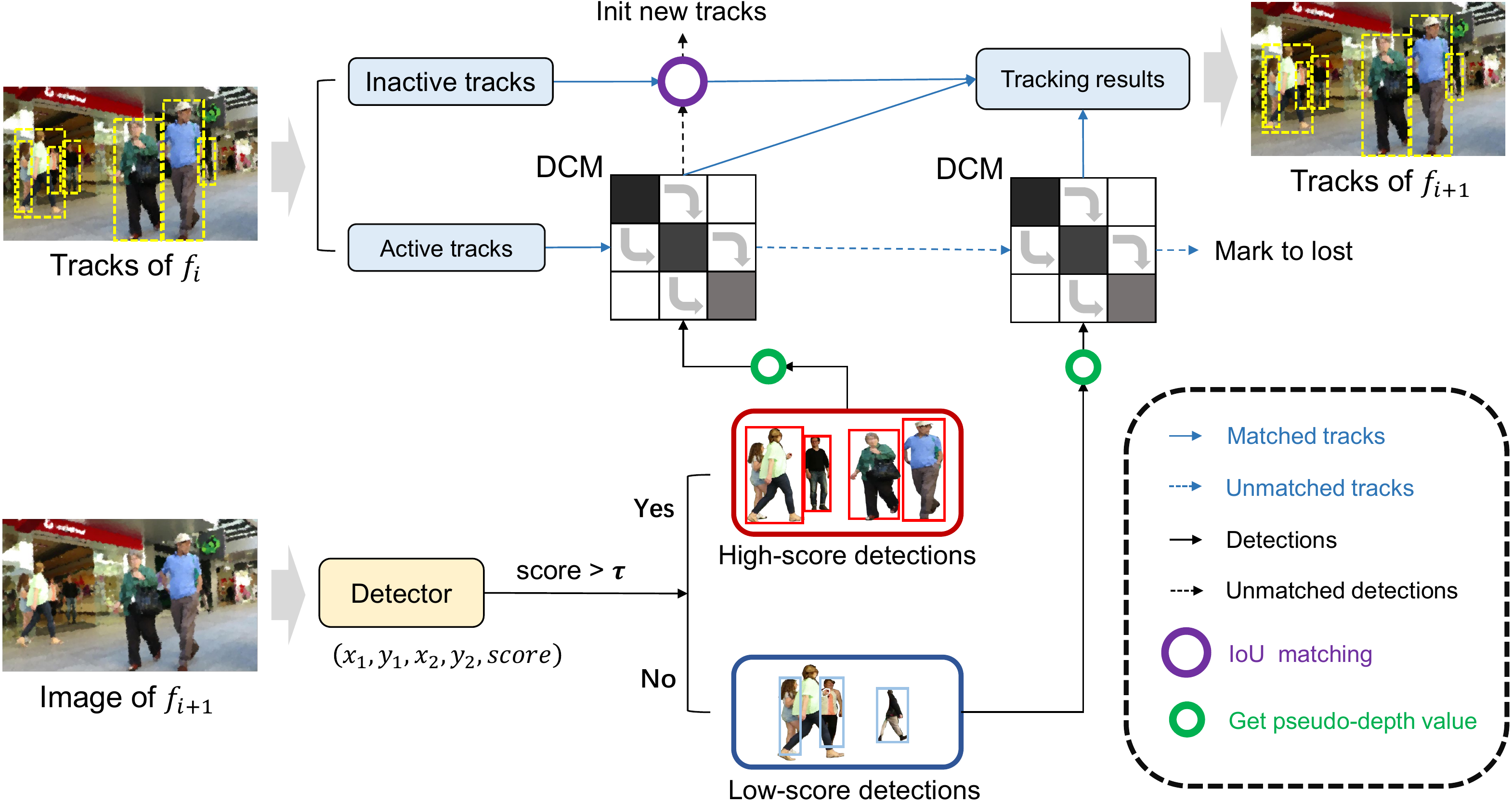}
\caption{
The overall framework of SparseTrack. The legend information is located in the box at the bottom right corner. $\tau$ is used to divide high-score and low-score detections. SparseTrack performs the target set decomposition to associate low-score occlusions accurately via DCM and pseudo-depth method.
}
\label{overview}
\end{figure*}

\begin{figure}[!t]
\centering
\includegraphics[width=0.97\linewidth, height=6cm]{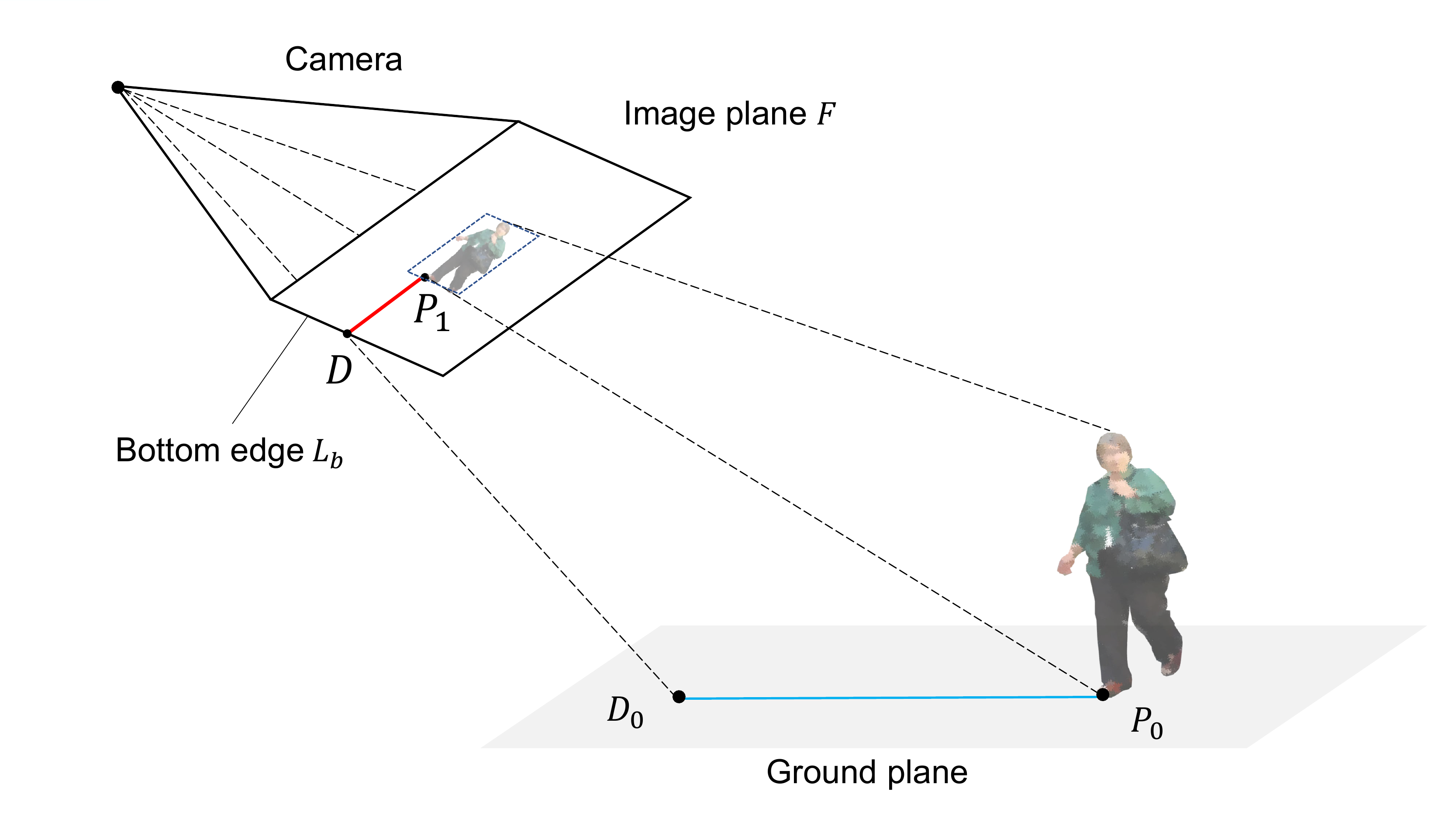}
\caption{The illustration of pseudo-depth method. The pseudo depth value is equal to the Euclidean distance between $P_1$ and $D$. In fact, pseudo-depth can be regarded as a projection that can be obtained by projecting the distance between $P_0$ and $D_0$ in 3D space onto the image plane. In the projection transformation, since the Euclidean distance between $P_0$ and $D_0$ is positively correlated with the pseudo-depth value, the pseudo-depth can reflect the depth of the target in the scene.}
\label{pdd}
\end{figure}

\subsection{General Framework}
SparseTrack follows the paradigm of tracking-by-detection and performs frame-by-frame data association. The overall framework is shown in~\cref{overview}. Given a frame $f_i$ in a video sequence, it is first processed by the YOLOX \cite{yolox} detector, which outputs detection results in the form of bounding boxes $(x_1, y_1, x_2, y_2)$ and confidence scores $s$. Then based on each detected bounding box and input image size, we use the pseudo-depth method to obtain pseudo-depth values of targets in the 2D image. The specific details are described in \cref{psd}. Pseudo-depth values can reflect the relative depth relationship between targets in the image. In the data association process, we utilize the pseudo-depth values of the detection and trajectory instances and use the DCM algorithm described in \cref{dccm} to accomplish the decomposition and association for the tracks and the detections, separately. Finally, SparseTrack outputs tracking results as $(x_1, y_1, w, h, s)$, where the coordinates $(w, h)$ represent the width and height of instance bounding boxes.

\subsection{Pseudo-Depth Method}\label{psd}
The \lzl{enabled} conditions of pseudo-depth method is based on two prior assumptions in general tracking scenes: the camera that captures objects is above the ground and all objects in the scene are on the same plane. These assumptions are easily satisfied in general scenarios. Specifically, pseudo-depth method is a simple geometric method for obtaining pseudo-depth values. Firstly, we acquire the position point $P_0$ where the object in 3D space contacts the ground, project point $P_0$ onto the image plane $F$ of the camera, and obtain the corresponding projection point $P_1$. Secondly, we draw a perpendicular from point $P_1$ to the bottom edge $L_b$ of the image and obtain the intersection point $D$ between the perpendicular and the bottom edge $L_b$. We define the Euclidean distance between the projection point $P_1$ and the intersection point $D$ as the pseudo-depth value of objects, as shown in ~\cref{pdd}. \lzl{Set the height of the image to $H$}, The pixel coordinates of $P_1$ is denoted as $(x_p, y_p)$. The pseudo-depth is computed as: 
\begin{equation}
\label{e-pdd}
\lzl{L_p = H - y_p},
\end{equation}
where $L_p$ is the pseudo-depth value. Although pseudo-depth values can not represent the ground truth depth of the target in 3D space, it is able to sufficiently measure the relative depth relationship between different objects on the same ground, which also provides a basis for dense crowd division at the depth level.

\subsection{Depth Cascade Matching}\label{dccm}
With the depth information obtained by the pseudo-depth method, we can divide targets in the scene based on their pseudo-depth values, resulting in a sparser distribution of previously dense crowds. Firstly, we compute the minimum pseudo-depth value of the detection set $D_{d\text{-}min}$ and the maximum pseudo-depth value of the detection set $D_{d\text{-}max}$. Then we uniformly divide the interval distance between $D_{d\text{-}min}$ and $D_{d\text{-}max}$ into $k$ depth intervals $\{I_0, I_1, ..., I_{k-1}\}_{det}$, which serve as $k$ different depth levels. With the same method, we can also obtain the minimum pseudo-depth value of the trajectory set $D_{t\text{-}min}$, maximum pseudo-depth value of the trajectory set $D_{t\text{-}max}$, and $k$ depth intervals $\{I_0, I_1, ..., I_{k-1}\}_{track}$. Based on the detection depth intervals $\{I_0, I_1, ..., I_{k-1}\}_{det}$ and the track depth intervals $\{I_0, I_1, ..., I_{k-1}\}_{track}$, we separately obtain the trajectory subsets $T_{sub}$ and detection subsets $D_{sub}$ that are located at different depth levels according to pseudo-depth values. Subsequently, the tracker performs IoU association $\mathcal{I}$ between the trajectory and detection subsets at the same depth level. Unmatched trajectories $\mathcal{T}_0$ and detections $\mathcal{D}_0$ will participate in the association process of the next depth level after the data association at each depth level. The pseudo-code of the depth cascade matching algorithm is shown in~\cref{algo:dcm}. 

\begin{algorithm}[h]
\SetAlgoLined
\DontPrintSemicolon
\SetNoFillComment
\footnotesize
\KwIn{tracks set $\texttt{T}$; detections set $\texttt{D}$; the number of depth levels $\texttt{k}$; Hungarian matching $\mathcal{H}$; the function of IoU distance metric $\mathcal{I}$; the function of getting max pseudo-depth value $\mathcal{M}_{max}$; the function of getting min pseudo-depth value $\mathcal{M}_{min}$; initialize matching threshold $\tau$}
\KwOut{Matched tracks $\mathcal{T}$; Unmatched tracks $\mathcal{T}_0$; Umatched detections $\mathcal{D}_0$}
Initialization: $\mathcal{T} \leftarrow \emptyset$, $\mathcal{T}_0 \leftarrow \emptyset$, $\mathcal{D}_0 \leftarrow \emptyset$\;
\BlankLine
\tcc{generate sparse subsets of detections}
$D_{d\text{-}max} \leftarrow \mathcal{M}_{max}(\texttt{D})$\;
\BlankLine
$D_{d\text{-}min} \leftarrow \mathcal{M}_{min}(\texttt{D})$\;
\BlankLine
$\{I_0,...,I_{k-1}\}_{det} \leftarrow split([D_{d\text{-}min},  D_{d\text{-}max}], k)$\;
\BlankLine
$D_{sub} \leftarrow \emptyset$  \tcc{all of subsets of detections} 
\BlankLine
\For{ $\{I_i\}_{det}$ in $\{I_0,...,I_{k-1}\}_{det}$ }{
    \BlankLine
    $D_i \leftarrow \emptyset$\;
    \BlankLine
    \For{$d$ in $\texttt{D}$}{
        \BlankLine
        \If{$d.depth$ $\in$ $\{I_i\}_{det}$}{
            \BlankLine
           $D_i \leftarrow D_i \cup d$\;
        }
    }
    $D_{sub} \leftarrow  D_{sub} \cup D_i$\;
}

\BlankLine
\tcc{generate sparse subsets of tracks}
$D_{t\text{-}max} \leftarrow \mathcal{M}_{max}(\texttt{T})$\;
\BlankLine
$D_{t\text{-}min} \leftarrow \mathcal{M}_{min}(\texttt{T})$\;
\BlankLine
$\{I_0,...,I_{k-1}\}_{track} \leftarrow split([D_{t\text{-}min},  D_{t\text{-}max}], k)$\;
\BlankLine
$T_{sub} \leftarrow \emptyset$ \tcc{all of subsets of tracks} 
\BlankLine
\For{$\{I_i\}_{track}$ in $\{I_0,...,I_{k-1}\}_{track}$}{
    \BlankLine
    $T_i \leftarrow \emptyset$\;
    \BlankLine
    \For{$t$ in $\texttt{T}$}{
        \BlankLine
        \If{$t.depth$ $\in$ $\{I_i\}_{track}$}{
            \BlankLine
           $T_i \leftarrow T_i \cup t$\;
        }
    }
    $T_{sub} \leftarrow  T_{sub} \cup T_i$\;
}
    
\BlankLine
\tcc{depth cascade matching}
\For{track subset $T_i$, detection subset $D_i$ in ($T_{sub}, D_{sub}$)}{
    \BlankLine
    \tcc{unmatched objects from previous stage participate the association of current stage.}
    \BlankLine
    $T_i \leftarrow T_i \cup \mathcal{T}_0$\;
    \BlankLine
    $D_i \leftarrow D_i \cup \mathcal{D}_0$\;
    \BlankLine
    
    \tcc{get cost matrix based on IoU distance}
    \BlankLine
    $C_i \leftarrow \mathcal{I}(T_i, D_i)$\;
    \BlankLine
    
    \tcc{association}
    \BlankLine
    $T_{matched}, \mathcal{T}_0, \mathcal{D}_0 \leftarrow \mathcal{H}(C_i, \tau)$\;
    \BlankLine
    
    \tcc{add matched tracks}
    \BlankLine
    $\mathcal{T} \leftarrow  \mathcal{T} \cup T_{matched}$\;
}
Return: $\mathcal{T}$, $\mathcal{T}_0$, $\mathcal{D}_0$
\caption{Depth Cascade Matching.}
\label{algo:dcm}
\end{algorithm}

\subsection{The Association Process of SparseTrack}
We propose a simple yet powerful data association method. Unlike other methods \cite{strongsort, deepsort, centertrack, TraDeS, CStrack, qdtrack, fairmot} that focus on improving the appearance features and motion cues to ensure the reliability of association, SparseTrack divides all targets in the scene into multiple sparse target subsets by the pseudo-depth method and DCM to effectively associate occlusions in crowds. \lzl{We divide the detection set into high-score and low-score detection subsets based on confidence scores via~\cite{bytetrack}}. Then, we use the aforementioned depth cascaded matching algorithm to perform data association on the high-score detection subset. For trajectories that appear in the previous frame but are not matched, we associate them with the low-score detection subset via the depth cascade matching algorithm. For unmatched high-score detections, we associate it with newly appeared trajectories in the previous frame. Finally, unmatched high-score detections are initialized as new trajectories, while lost trajectories that exceed the max number of losing frames are removed. The pseudo-code of data association of SparseTrack is shown in~\cref{algo:sptrack}.

\begin{algorithm}[t]
\SetAlgoLined
\DontPrintSemicolon
\SetNoFillComment
\footnotesize
\KwIn{A video sequence $\texttt{V}$; object detector $\texttt{Det}$; high-score detection  threshold $\tau$; the function of depth cascade matching $\texttt{DCM}$; the Kalman motion model $\texttt{KF}$}
\KwOut{tracking results of the video $\mathcal{T}$}
Initialization: $\mathcal{T} \leftarrow \emptyset$\;
\For{frame $f_i$ in \texttt{V}}{
    \BlankLine
    \tcc{get high-score and low-score detections from current frame}
    \BlankLine
    ${D}_i \leftarrow \texttt{Det}(f_i)$  \tcc{detections per frame}
    \BlankLine
    ${D}_{high} \leftarrow \emptyset$\;
    \BlankLine
    ${D}_{low} \leftarrow \emptyset$\;
    \BlankLine
    \BlankLine
    
    \For{$d$ in ${D}_i$}{
        \If{$d_{score} > \tau $}{
            \BlankLine
            ${D}_{high} \leftarrow {D}_{high} \cup \{d\}$\;
        }
        \Else{
            \textcolor{codegreen}{${D}_{low} \leftarrow {D}_{low} \cup \{d\}$\;}
        }
    }
    \BlankLine
    \BlankLine
    
    \tcc{predict the location of tracks from previous frames}
    \BlankLine
    $\mathcal{T} \leftarrow \texttt{KF}(\mathcal{T})$\;
    \BlankLine
    \BlankLine
    
    \tcc{associate high-score detections by $\texttt{DCM}$}
    \BlankLine
    \textcolor{codegreen}{${T}_{matched},  {D}_{unmatched} \leftarrow \texttt{DCM}(\mathcal{T}, D_{high})$\;}
    \BlankLine
    ${T}_{unmatched} \leftarrow \emptyset$\;
    \BlankLine
    \For{$t$ in $(\mathcal{T} - {T}_{matched})$}{
        \If{$t_{state}$ is not lost}{
            \BlankLine
            ${T}_{unmatched} \leftarrow {T}_{unmatched} \cup \{t\}$
        }
    }
    \BlankLine
    \BlankLine
    
    \tcc{associate low-score detections by $\texttt{DCM}$}
    \BlankLine
    \textcolor{codegreen}{${T}_{re-matched} \leftarrow \texttt{DCM}({T}_{unmatched}, D_{low})$\;}
    \BlankLine
    \BlankLine
    
    \tcc{update tracking results}
    \BlankLine
    $\mathcal{T} \leftarrow \{ {T}_{matched}, {T}_{re-matched}\}$\;
    \BlankLine
    \BlankLine
    
    \tcc{add new tracks}
    \BlankLine
    \For{$d$ in $\mathcal{D}_{unmatched}$}{
        \BlankLine
        $\mathcal{T} \leftarrow \mathcal{T} \cup \{d\}$
    }
}
Return: $\mathcal{T}$
\caption{The Data Association of SparseTrack.}
\algorithmfootnote{Track confirmation is not shown in the pseudo-code for simplicity. Function $\texttt{DCM}$ can refer to~\cref{algo:dcm}. The key steps of SparseTrack are in \textcolor{codegreen}{green}.}
\label{algo:sptrack}
\end{algorithm}

As the confidence score is related to the occlusion of the target, we inherit the decomposition method based on the confidence score from \cite{bytetrack}. Since high-score detections often mean a lower degree of occlusion, we set the number of pseudo-depth levels to 1 or 2 during associating high-score objects. Conversely, for associating low-score detections, we set the number of pseudo-depth levels to 4 or 8, which helps in the fine-grained association of dense occlusions. It is worth mentioning that the DCM is plug-and-play and can be integrated into various trackers. The specific settings are detailed in \cref{exp}. 

\section{Experiments}\label{exp}
\subsection{Setting}
\subsubsection{Datasets}We evaluate SparseTrack on the MOT17 \cite{mot16}, MOT20 \cite{mot20}, and DanceTrack \cite{dancetrack} benchmark datasets, and submit the evaluation results to compare the performance with other methods. For ablation experiments, we split half of the training sets of MOT17 and MOT20 (dense scene) as validation sets \lzl{and conduct experiments on the MOT17 validation set, MOT20 validation set and DanceTrack validation set to analyse the effects with different settings}.
\subsubsection{Metrics}We use the common CLEAR metrics \cite{CLEAR-metrics} (MOTA, FP, FN, IDs, \etc), IDF1 \cite{idf1}, HOTA \cite{hota}, AssA and DetA, to evaluate tracking performance. MOTA is computed based on FP, FN, and IDs. As the number of FP and FN is larger than that of IDs, MOTA prefers to reflect detection performance. Additionally, IDF1 and AssA focus more on association performance. DetA is used to measure detection accuracy. HOTA is a comprehensive measure used to evaluate the overall effectiveness of detection and association.
\subsubsection{Implementation details}During the data association stage, we set different numbers of pseudo-depth levels for different datasets according to the scene crowding level. For MOT17, we set the default number of pseudo-depth levels to 3 and keep the maximum length of lost tracks at 30 frames. For MOT20, we set the default number of depth levels to 8 and keep the maximum length of lost tracks at 60 frames. For DanceTrack, we set the number of depth levels to 12 and keep the lost tracks until 60 frames. \lzl{The high-score and low-score detection threshold is set to 0.6 and 0.1 by default, separately, which aligns with the baseline approach}. To align the positions of detections and tracks on depth levels as closely as possible, we implement a fast online version of global motion compensation (GMC). It is worth noting that all our experimental results rely on IoU distance association and do not use any appearance feature component.

\lzl{During the inference phase, we configure the input resolution as $800\times1440$ for MOT17 and DanceTrack, while we utilize an input resolution of $896\times1600$ for MOT20. To make a fair comparison with the baseline method \cite{bytetrack}, we adopt the pre-trained YOLOX \cite{yolox} detector from ByteTrack and use the same weights and non-maximal suppression (NMS) thresh as the baseline method. For the ablation experiments on the MOT20 validation set, we train SparseTrack on the CrowdHuman and the half of MOT20 train set. For the ablation experiments on the DanceTrack validation set, we adopt publicly available YOLOX detector from \cite{dancetrack}. It is worth mentioning that DanceTrack only provides pre-trained detectors of the YOLOX-x model.} All implementations are performed on NVIDIA GeForce RTX 3090 GPUs.
 
\subsection{Evaluation of Different Benchmark}
We evaluate SparseTrack on the test set of MOT17, MOT20 and DanceTrack to compare with other methods. All evaluation results are based on the private detection protocol and are shown in \cref{tab:17}, \cref{tab:20} and \cref{tab:dance}, respectively \footnote{The best results are shown with bold in \cref{tab:17}, \cref{tab:20} and \cref{tab:dance}. In \cref{tab:dance}, \textbf{$*$} indicates that the tracker utilize extra training data.}.

\begin{table}[t]
\caption{The comparison of SparseTrack with other methods on the MOT17 test set.}
\label{tab:17}
\resizebox{1.0\linewidth}{!}{
\setlength{\tabcolsep}{1.73pt}
 
\begin{tabular}{ l | c c c c c c r}
\toprule
Tracker & HOTA$\uparrow$  & MOTA$\uparrow$ & IDF1$\uparrow$ &  FP$\downarrow$ & FN$\downarrow$ & IDs$\downarrow$ & FPS$\uparrow$\\
\midrule
\textcolor{red}{\textbf{$enhance$ $motion$ :}}\\
Tube\_TK\cite{tubetk} & 48.0 & 63.0 & 58.6 & 27060 & 177483 & 4137 & 3.0\\
CenterTrack\cite{centertrack} &52.2 &  67.8 & 64.7 & 18498 & 160332 & 3039 & 17.5\\
TraDes\cite{TraDeS}& 52.7  & 69.1 & 63.9 & 20892 & 150060 & 3555 & 17.5\\
MAT\cite{mat} & 53.8 & 69.5 & 63.1 & 30660 & 138741 & 2844 & 9.0\\
PermaTrackPr\cite{PermaTrackPr}  & 55.5 & 73.8 & 68.9& 28998 & 115104 & 3699 & 11.9\\
OC-SORT\cite{oc-sort} & 63.2 & 78.0 & 77.5 & \textbf{15129} &	107055 & 1950 & 29.0\\ 
MotionTrack\cite{motiontrack} & \textbf{65.1} & 81.1 & 80.1 & 23802 &  \textbf{81660} & \textbf{1140} & 15.7 \\

\midrule
\textcolor{red}{\textbf{$embedding$ :}}\\
DAN\cite{DAN} & 39.3 & 52.4 & 49.5 & 25423 & 234592 & 8431 & \textless 3.9\\
QuasiDense\cite{qdtrack} & 53.9 & 68.7 & 66.3 & 26589 & 146643 & 3378 & 20.3\\
SOTMOT\cite{SOTMOT} & - & 71.0 & 71.9 & 39537 & 118983 & 5184 & 16.0\\
Semi-TCL\cite{semi} & 59.8 & 73.3 & 73.2 & 22944 & 124980 & 2790 & -\\
FairMOT\cite{fairmot} & 59.3 & 73.7 & 72.3 & 27507 & 117477 & 3303 & 25.9\\
CSTrack\cite{CStrack} & 59.3 & 74.9 & 72.6 & 23847 & 114303 & 3567 & 15.8\\
SiamMOT\cite{siammot} & - & 76.3 & 72.3 & - & - & - & 12.8\\
ReMOT\cite{remot} & 59.7 & 77.0 & 72.0 & 33204 & 93612 & 2853 & 1.8\\
StrongSORT\cite{strongsort} & 64.4 & 79.6 & 79.5 & 27876 & 86205 & 1194 & 7.1\\
BoT-SORT-ReID\cite{BoT-SORT} & 65.0 & 80.5 & \textbf{80.2} & 22521 & 86037 & 1212 & 4.5\\

\midrule
\textcolor{red}{\textbf{$attention$ :}}\\
CTracker\cite{ctracker}  & 49.0 & 66.6 & 57.4& 22284 & 160491 & 5529 & 6.8\\
TransCenter\cite{transcenter} & 54.5 & 73.2 & 62.2 & 23112 & 123738 & 4614 & 1.0\\
RelationTrack\cite{relationtrack} & 61.0 & 73.8 & 74.7 & 27999 & 118623 & 1374 & 8.5\\
TransTrack\cite{transtrack} & 54.1 & 75.2 & 63.5 & 50157 & 86442 & 3603 & 10.0\\
MOTRv2\cite{motrv2} & 62.0 & 78.6 & 75.0 & 23409 & 94797 & 2619 & 7.5\\
P3AFormer\cite{p3aformer} & - & \textbf{81.2} & 78.1 & 17281 & 86861 &  1893 & -\\

\midrule
\textcolor{red}{\textbf{$graphical$ \& }}\\
\textcolor{red}{\textbf{$correlation$ :}}\\
GSDT\cite{GSDT} & 55.2 & 73.2 & 66.5 & 26397 & 120666 & 3891 & 4.9\\
FUFET\cite{FUFET} & 57.9 & 76.2 & 68.0 & 32796 & 98475 & 3237 & 6.8\\
CorrTracker\cite{CorrTrack}& 60.7  & 76.5 & 73.6 & 29808 & 99510 & 3369 & 15.6\\
TransMOT\cite{transmot} & 61.7 & 76.7 & 75.1 & 36231 & 93150 & 2346 & 9.6\\

\midrule
\textcolor{red}{\textbf{$IoU$ $only$ :}}\\
ByteTrack\cite{bytetrack} & 63.1 & 80.3 & 77.3 & 25491 & 83721 & 2196 & \textbf{29.6}\\
BoT-SORT\cite{BoT-SORT} & 64.6 & 80.6 & 79.5 & 22524 & 85398 & 1257 & 6.6\\
\textbf{SparseTrack (ours)} & \textbf{65.1} & 81.0 & 80.1 & 23904 & 81927 & 1170 & 19.9\\
\bottomrule
\end{tabular}
}
\end{table}

\textbf{MOT17.} By performing the target set decomposition based on pseudo-depth, SparseTrack achieves impressive performance on the MOT17 test set. Compared to the baseline (ByteTrack) with the \lzl{same pre-trained detector}, SparseTrack achieves gains of \textbf{+0.7} MOTA, \textbf{+2.8} IDF1, \textbf{+2.0} HOTA and IDs decrease to almost twice the original level. Instead of other methods that use appearance features and graph convolutional networks, SparseTrack achieves comparable performance with SOTA using only simple IoU association, which demonstrates that SparseTrack is \lzl{simple} and strong.

\begin{table}[t]
\caption{The comparison of SparseTrack with other methods on the MOT20 test set.}
\label{tab:20}
\setlength{\tabcolsep}{1.0pt}
\resizebox{1.0\linewidth}{!}{
\begin{tabular}{ l | c c c c c c r}
\toprule
Tracker & HOTA$\uparrow$ & MOTA$\uparrow$ & IDF1$\uparrow$ & FP$\downarrow$ & FN$\downarrow$ & IDs$\downarrow$ & FPS$\uparrow$\\
\midrule
\textcolor{red}{\textbf{$enhance$ $motion$ :}}\\
MotionTrack\cite{motiontrack}  & 62.8 & 78.0 & 76.5& 28629 & 84152 & 1165 & 15.0\\

\midrule
\textcolor{red}{\textbf{$embedding$ :}}\\
FairMOT\cite{fairmot}  & 54.6 & 61.8 & 67.3 & 103440 & 88901 & 5243 & 13.2\\
Semi-TCL\cite{semi}  & 55.3 & 65.2 & 70.1 & 61209 & 114709 & 4139 & -\\
CSTrack\cite{CStrack}  & 54.0 & 66.6 & 68.6 & 25404 & 144358 & 3196 & 4.5\\
SiamMOT\cite{siammot}  & - & 67.1 & 69.1 & - & - & - & 4.3\\
SOTMOT\cite{SOTMOT}  & - & 68.6 & 71.4 & 57064 & 101154 & 4209 & 8.5\\
BoT-SORT-ReID\cite{BoT-SORT} & 63.3 & 77.8 & \textbf{77.5} & 24638 & 88863 & 1257 & 2.4\\
UTM\cite{UTM} & 62.5 & \textbf{78.2} & 76.9 & 29964 & \textbf{81516} & 1228 & -\\

\midrule
\textcolor{red}{\textbf{$attention$ :}}\\
TransCenter\cite{transcenter}   & - & 61.9 & 50.4& 45895 & 146347 & 4653 & 1.0\\
TransTrack\cite{transtrack}   & 48.5 & 65.0 & 59.4& 27197 & 150197 & 3608 & 7.2\\
RelationTrack\cite{relationtrack}  & 56.5 & 67.2 & 70.5 & 61134 & 104597 & 4243 & 2.7\\
MOTR\cite{motr} & 57.8  & 73.4 & 68.6 & - & - & 2439 & -\\
P3AFormer\cite{p3aformer} & - & 78.1 & 76.4 &  25413   & 86510 & 1332 & -\\

\midrule
\textcolor{red}{\textbf{$graphical$ \& }}\\
\textcolor{red}{\textbf{$correlation$ :}}\\
MLT\cite{MLT}  & 43.2 & 48.9 & 54.6 & 45660 & 216803 & 2187 & 3.7\\
CorrTracker\cite{CorrTrack}  & - & 65.2 & 69.1 & 79429 & 95855 & 5183 & 8.5\\
GSDT\cite{GSDT}  & 53.6 & 67.1 & 67.5 & 31913 & 135409 & 3131 & 0.9\\

\midrule
\textcolor{red}{\textbf{$IoU$ $only$ :}}\\
ByteTrack\cite{bytetrack} & 61.3 & 77.8 & 75.2 & 26249 & 87594 & 1223 & \textbf{17.5}\\
BoT-SORT\cite{BoT-SORT} & 62.6 & 77.7 & 76.3 & \textbf{22521} & 86037 & 1212 & 6.6 \\
\textbf{SparseTrack (ours)} & \textbf{63.4} & \textbf{78.2} & 77.3 & 25108 & 86720 & \textbf{1116} & 12.5\\
\bottomrule
\end{tabular}
}
\end{table}

\textbf{MOT20.} Compared to MOT17, MOT20 has denser scenes, more occlusions, and longer videos, which increases the difficulty of the tracker in handling occlusions. SparseTrack uses the same \lzl{pre-trained detector} as the baseline and achieves a gain of \textbf{+0.4} MOTA, \textbf{+2.1} IDF1, and \textbf{+2.1} HOTA. Compared to other methods that utilize appearance cues, attention mechanism \cite{attention, vit} and graph \cite{graph-mot,graph-mot1,Learnable_graph_matching}, SparseTrack provides a more convenient solution: the target set decomposition based on hierarchical depth. It is worth mentioning that SparseTrack achieves very low IDs and FP, as well as high HOTA, which indicates that the target set decomposition is helpful in associating dense occlusions.

\begin{table}[t]
\caption{The comparison of SparseTrack with other methods on the DanceTrack test set.}
\label{tab:dance}
\centering
\setlength{\tabcolsep}{1.0pt}
\begin{tabular}{ l | c c c c r}
\toprule
Tracker & HOTA$\uparrow$ & MOTA$\uparrow$ & IDF1$\uparrow$ & AssA$\uparrow$ & DetA$\uparrow$ \\
\midrule
\textcolor{red}{\textbf{$enhance$ $motion$ :}}\\
CenterTrack\cite{centertrack} & 41.8 &  86.8 & 35.7 & 22.6 & 78.1\\
TraDes\cite{TraDeS}& 43.3 & 86.2 & 41.2 & 25.4 & 74.5\\
OC-SORT\cite{oc-sort} & 55.1 & \textbf{92.0} & 54.6 & 38.3 & \textbf{80.3}\\ 

\midrule
\textcolor{red}{\textbf{$embedding$ :}}\\
FairMOT\cite{fairmot}  & 39.7 & 82.2 & 40.8 & 23.8 & 66.7 \\
FCG$^*$\cite{FCG} & 48.7 & 89.9 & 46.5 & 29.9 & 79.8 \\
QuasiDense\cite{qdtrack} & 54.2 & 87.7 & 50.4 & 36.8 & 80.1\\

\midrule
\textcolor{red}{\textbf{$attention$ :}}\\
TransTrack\cite{transtrack}   & 45.5 & 88.4 & 45.2 & 27.5 & 75.9 \\
GTR\cite{GTR} & 48.0 & 84.7 & 50.3 & 31.9 & 72.5\\
MOTR\cite{motr} & 54.2  & 79.7 & 51.5 & \textbf{40.2}& 73.5 \\

\midrule
\textcolor{red}{\textbf{$IoU$ $only$ :}}\\
ByteTrack\cite{bytetrack} & 47.7 & 89.6 & 53.9 & 32.1 & 71.0 \\
BoT-SORT\cite{BoT-SORT} & 54.7 & 91.3 & 56.0 & 37.8 & 79.6 \\
\textbf{SparseTrack (ours)} & \textbf{55.5} & 91.3 & \textbf{58.3} & 39.1 & 78.9 \\
\bottomrule
\end{tabular}

\end{table}

\textbf{DanceTrack.} DanceTrack is a more challenging benchmark for multi-object tracking with frequent occlusions, non-rigid motions and similar appearances. \lzl{With the same pre-trained detector}, SparseTrack achieves significant improvements compared to the baseline with a gain of \textbf{+7.8} HOTA, \textbf{+1.7} MOTA, \textbf{+4.4} IDF1, \textbf{+7.0} AssA and \textbf{+7.9} DetA, which demonstrates the enormous potential of depth-based target set decomposition in handling occlusions. Even with simple IoU distance association, SparseTrack still achieves comparable or even better performance than other methods.

\subsection{Ablations}
\subsubsection{The impact of the number of pseudo-depth levels}
\lzl{As DCM focuses on addressing dense occlusions, we conduct experiments on the MOT20 validation set and DanceTrack validation set (More crowded scenario and more frequent occlusions.) to better observe the impact of the number of pseudo-depth levels on tracking performance. Specifically, we use the BYTE \cite{bytetrack} association method as the baseline and integrate DCM into the low-score association process within BYTE. We set the number of pseudo-depth levels for the process of associating low-score detections to 1, 2, 5, 7, 9, separately. It is worth noting that we lower the detection confidence score from the initial 0.1 to 0.01 to involve more low-scoring detections in the association process.} The experimental results are shown in \cref{ablation:levels-compare}. 

\lzl{After integrating DCM, the association performance of BYTE gradually improves with the increase in the number of pseudo-depth levels. It indicates that a higher number of pseudo-depth levels is more beneficial to associate low-score occlusions. However, it does not yield additional gains when the number of pseudo-depth levels is exorbitant (\eg the number of pseudo-depth levels exceeds 7). We speculate that excessively frequent layering schemes result in increased sparsity among both low-scoring detections and participating tracks, which can not yield additional benefits compared to an already sparse scheme.}
\begin{table}[t]
\caption{
\lzl{Comparison of the association performance with different numbers of the pseudo-depth level on MOT20 and DanceTrack validation sets.}
}
\label{ablation:levels-compare}
\renewcommand\arraystretch{1.1}
\centering
\setlength{\tabcolsep}{1.0pt}
\begin{tabular}{ c c |c c c |c c c}
\toprule
\multirow{2.5}{*}{Levels} & &\multicolumn{3}{c}{MOT20}  & \multicolumn{3}{c}{DanceTrack} \\
\cmidrule(r){3-5} \cmidrule(lr){6-8}
 & & HOTA$\uparrow$ & AssA$\uparrow$  & IDF1$\uparrow$ & HOTA$\uparrow$ & AssA$\uparrow$ & IDF1$\uparrow$ \\
\midrule
1 & & 68.6 & 65.9 & 83.0 & 52.6 & 36.5 & 54.2\\
2 & & 68.7 & 66.1 & 83.1 & 53.4 & 37.7 & 55.4\\
5 & & 68.8 & 66.1 & 83.2 & \textbf{53.9} & \textbf{38.5} & \textbf{55.8}\\
7 & & \textbf{68.9} & \textbf{66.3} & \textbf{83.4} & 53.8 & 38.4 & 55.7\\
9 & & 68.9 & 66.3 & 83.3 & 53.8 & 38.3 & 55.7\\
\bottomrule
\end{tabular}

\end{table}

\begin{figure}[!t]
\centering
\includegraphics[width=0.99\linewidth, height=10cm]{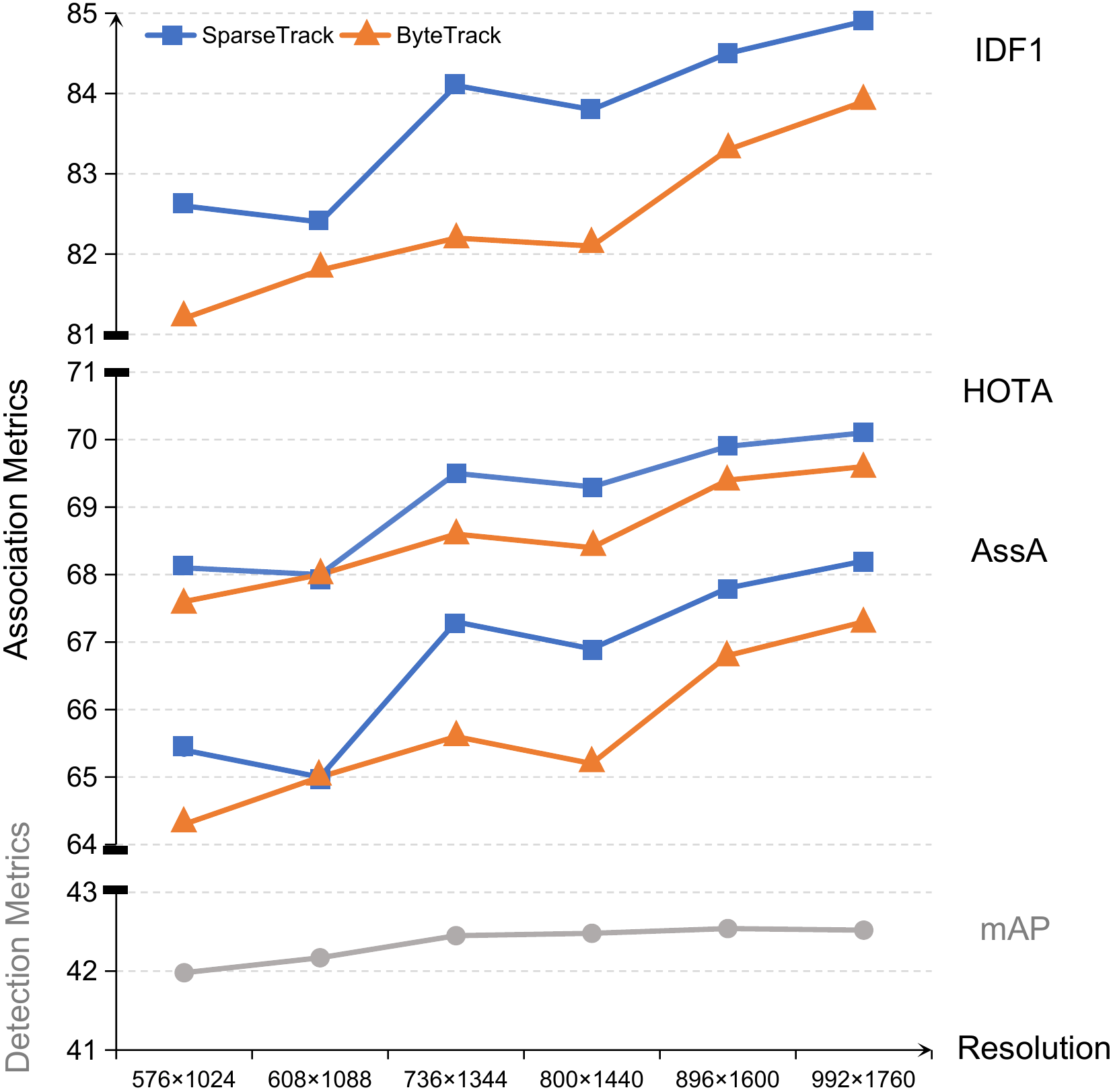}
\caption{\lzl{The graph depicts the changes in the association performance of SparseTrack and ByteTrack \cite{bytetrack} as the resolution varies. While there is marginal improvement in detector performance with increasing resolution, the association performance of the trackers consistently strengthens. It is worth noting that SparseTrack demonstrates more significant enhancements compared to ByteTrack.}}
\label{resolution-compare}
\end{figure}

\subsubsection{The impact of the input with different resolutions}
\lzl{We investigate the impact of input resolution on the tracking performance of SparseTrack. Specifically, we maintain consistent detection settings and test the association performance of SparseTrack and ByteTrack on the MOT20 validation dataset with varying input resolutions. We set the pseudo-depth levels to 8 in SparseTrack for associating low-scoring detections. The results are shown in} \cref{resolution-compare}.

\lzl{As the increase in resolution, the performance of object detection shows a gradual improvement. However, beyond a certain threshold in resolution, the detector exhibits diminishing returns, with little to no substantial gains. Interestingly, both SparseTrack and ByteTrack show significant enhancements in their tracking performance as resolution continues to climb. This phenomenon can be attributed to the increased detection capability to recognize and involve more severely occluded targets within the tracking process. Notably, our observations suggest that SparseTrack outperforms ByteTrack in terms of association enhancements during the process.}

\subsubsection{The impact of the global motion compensation}
\lzl{We perform ablative analyses of the global motion compensation (GMC) module using SparseTrack on both the MOT17 and MOT20 validation datasets. During the low-score association stage, we fix the number of pseudo-depth levels at 8. We maintain the default input resolution settings for MOT17 and MOT20. The results are presented in} \cref{ablation:gmc-compare}.

\lzl{By incorporating GMC, substantial enhancements in the performance of SparseTrack are achieved on the MOT17 validation dataset. This improvement primarily stem from the more pronounced camera motion prevalent in the MOT17 dataset. Conversely, GMC yield little gains in tracking performance on the MOT20 dataset.}
\begin{table}[!t]
\caption{
\lzl{Comparison of the association performance with or without global motion compensation on MOT17 and MOT20 validation sets.}
}
\label{ablation:gmc-compare}
\renewcommand\arraystretch{1.1}
\centering
\setlength{\tabcolsep}{1.0pt}
\begin{tabular}{c c |c c c |c c c}
\toprule
\multirow{2.55}{*}{w/ GMC}& & \multicolumn{3}{c}{MOT17}  & \multicolumn{3}{c}{MOT20} \\
\cmidrule(r){3-5} \cmidrule(lr){6-8}
 & & HOTA$\uparrow$ & AssA$\uparrow$  & IDF1$\uparrow$ & HOTA$\uparrow$ & AssA$\uparrow$ & IDF1$\uparrow$ \\
\midrule
           & & 67.9 & 69.9 & 79.1 & 69.9 & 67.8 & 84.5\\
\checkmark & & 69.2 & 72.3 & 81.4 & 69.9 & 67.7 & 84.4\\
\bottomrule
\end{tabular}
\end{table}

\begin{table*}[t]
\caption{
\lzl{Comparison of the association methods on MOT17, MOT20 and DanceTrack validation sets.}
}
\label{methods-compare}
\renewcommand\arraystretch{1.1}
\centering
\setlength{\tabcolsep}{2.0pt}
\begin{tabular}{ c |c c c c |c c c c |c c c c}
\toprule
\multirow{2.5}{*}{Methods} & \multicolumn{4}{c}{MOT17}  & \multicolumn{4}{c}{MOT20} & \multicolumn{4}{c}{DanceTrack}\\
\cmidrule(r){2-5} \cmidrule(lr){6-9} \cmidrule(lr){10-13}
 & HOTA$\uparrow$ & AssA$\uparrow$  & IDF1$\uparrow$ & MOTA$\uparrow$ & HOTA$\uparrow$ &AssA$\uparrow$ & IDF1$\uparrow$ & MOTA$\uparrow$ & HOTA$\uparrow$ & AssA$\uparrow$ & IDF1$\uparrow$ & MOTA$\uparrow$ \\
\midrule
SORT \cite{Bewley2016_sort} &62.9 &61.9  &70.5  &73.5  &65.2 &59.1 &74.8 &85.1 &43.3  &26.0  &40.8  &86.0 \\
ByteTrack \cite{bytetrack}   &68.0 &69.9  &79.1  &76.3  &69.4 &66.9 &83.3 &86.1 &52.2&36.2&54.3&87.5\\
BoT-SORT \cite{BoT-SORT}    &68.9 &71.2  &80.7  &\textbf{76.9}  &69.5 &67.1 &83.7 &86.1 &53.4&37.6&\textbf{56.1} & \textbf{87.7}\\
OC-SORT \cite{oc-sort}     &66.3 &68.8  &77.5  &74.2  &68.0 &64.9 &82.0 &85.1 &51.2 &34.8 &51.1 &85.1 \\
SparseTrack &\textbf{69.2} &\textbf{72.3}  &\textbf{81.4}  &76.8  &\textbf{69.9} &\textbf{67.8} &\textbf{84.5} &\textbf{86.1} & \textbf{53.8} & \textbf{38.5} & 56.0 & 87.1 \\
\bottomrule
\end{tabular}
\end{table*}

\subsection{Comparison with Other Simple Methods}
\lzl{We compare SparseTrack with other location-dependent association methods on the MOT17, MOT20, and DanceTrack validation datasets respectively. The results as shown in \cref{methods-compare}. To ensure a fair comparison, we remove all specific hyper-parameters for each video to prevent any bias towards any method and all experiments maintain the same YOLOX-x detector model weights and NMS \cite{girshick14CVPR} hyper-parameters. For the SparseTrack, the number of pseudo-depth layers during the low-score association phase is consistently set to 8 and the low-score detection threshold remains at the default value of 0.1. It is worth mentioning that SparseTrack without GMC module reports the results on the MOT20 and DanceTrack validation set.}
 
\lzl{Compared to previous methods that rely on position information and Kalman filter~\cite{kf}, the association approach of SparseTrack achieves the best tracking performance on the aforementioned three validation sets.}

\subsection{Visualization}
\begin{figure*}[!t]
\centering
\includegraphics[width=0.97\linewidth, height=5cm]{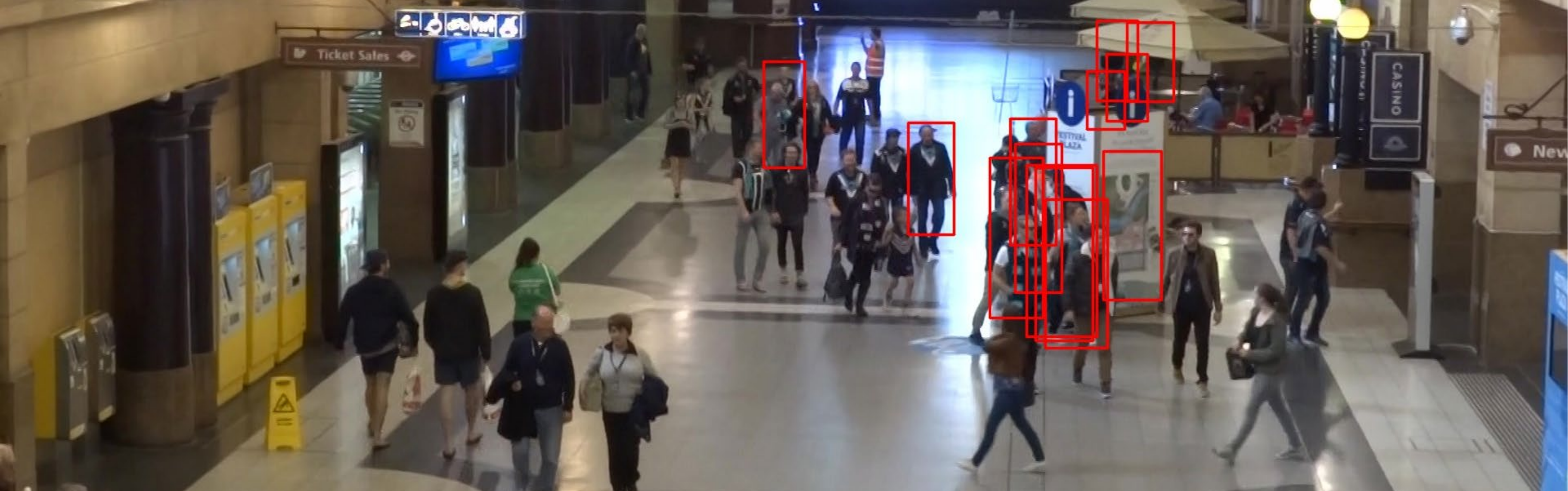}
\caption{Visualization of dense occlusions. We visualize ground truth instances with visibility less than 0.15. The occluded targets in the crowd are often clustered, resulting in dense occlusions.}
\label{dense-occ}
\end{figure*}

\begin{figure*}[!t]
\centering
\includegraphics[width=0.97\linewidth, height=4.3cm]{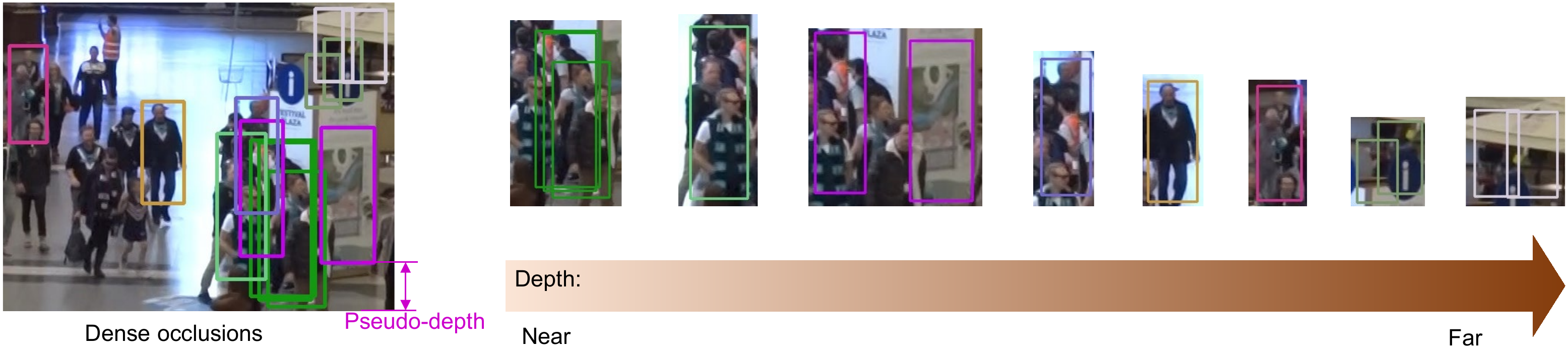}
\caption{Visualization of decomposing dense occlusions. We use different color of bounding boxes to distinguish different target subsets obtained via dividing pseudo-depth levels.}
\label{decomposition}
\end{figure*}

\begin{figure*}[!t]
\centering
\includegraphics[width=0.97\linewidth, height=24cm]{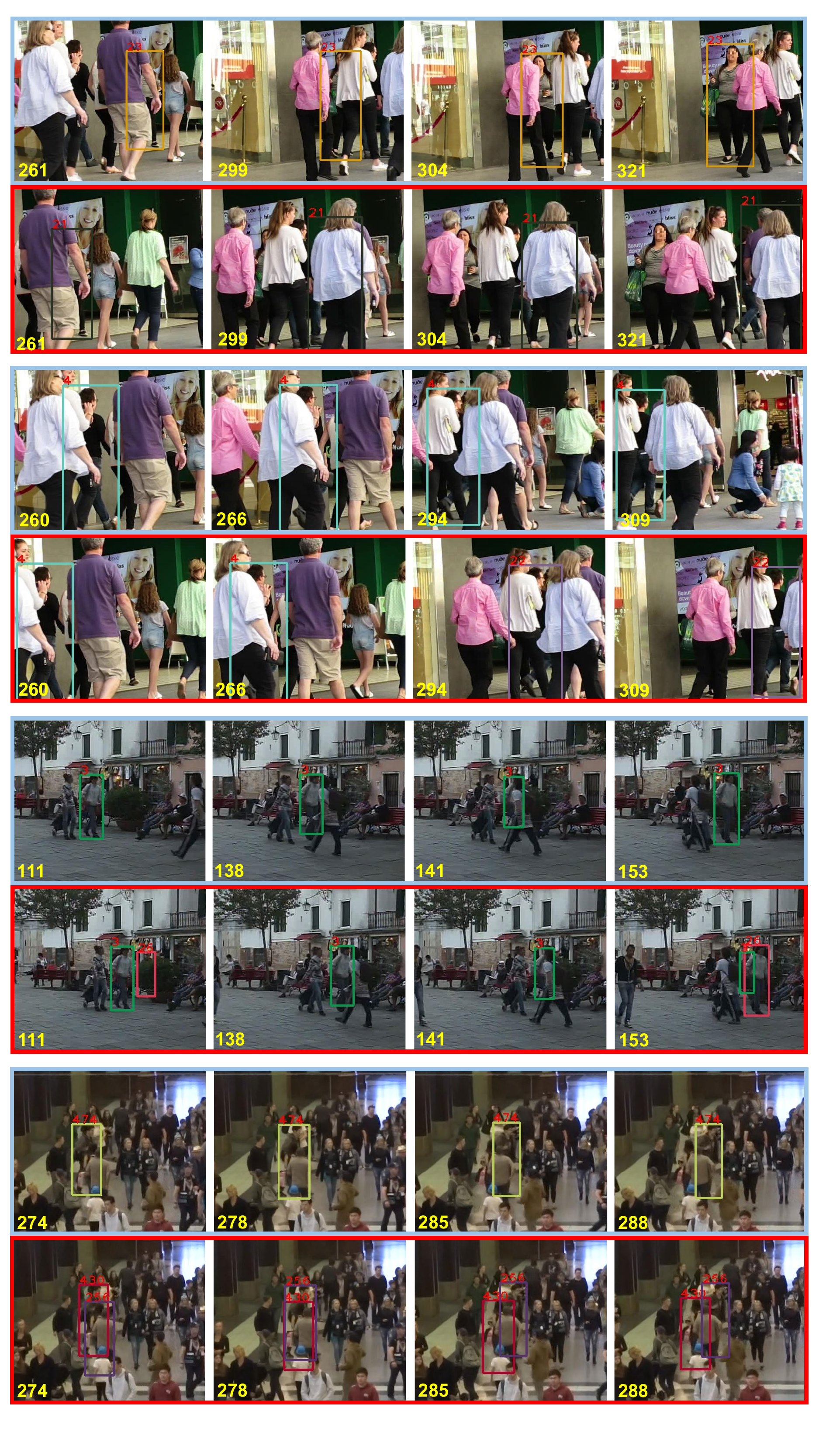}
\caption{\lzl{Comparable visualization of associating occlusive targets. We visualize the partial tracking results of SparseTrack and ByteTrack on the MOT17 and MOT20 data set, while keep the same settings of hyper-parameters on pre-trained detector. The tracking results of SparseTrack are displayed on the video clips in the} \textcolor{blue}{\textbf{blue box}}, \lzl{while the tracking results of ByteTrack are displayed on the video clips in the} \textcolor{red}{\textbf{red box}}. \lzl{The identical yellow numbers signify that the image cropping regions originate from the same frame.}}
\label{vis_occlusion}
\end{figure*}

To exhibiting the target set decomposition process and the performance of associating crowded targets intuitively, we conduct several different visualization experiments, as shown in \cref{dense-occ}, \cref{decomposition} and \cref{vis_occlusion}, respectively.

\textbf{Visualization of the target set decomposition.} 
Although the target set decomposition is a sub-process of DCM, it is the key to decoupling dense occlusions. In fact, not all targets in the scene are occluded state, but occluded targets may gather together and form dense occlusions easier, as shown in \cref{dense-occ}. By decomposing dense occlusion set based on pseudo-depth information, overlapping targets can be assigned to different depth levels, effectively. We visualize the decomposition of a set of dense occlusions, as shown in \cref{decomposition}.

\textbf{Visualization of associating occlusive targets via SparseTrack and ByteTrack.} 
\lzl{We conduct visualizations using both SparseTrack and ByteTrack on four different videos from the MOT challenge, while maintaining consistent detection hyper-parameter settings for the publicly available detector. The comparative results are shown in}~\cref{vis_occlusion}. \lzl{Comparing to ByteTrack, SparseTrack demonstrates greater stability in handling challenging occlusion scenarios. It is attributed to its ability to differentiate between targets at various depths crowded together, leveraging discriminative pseudo-depth information originating from the targets.}

\subsection{Discussion}
\lzl{While SparseTrack demonstrates remarkable performance improvements on both the MOT datasets and the DanceTrack dataset, it has poor performance on the public detection benchmark of the MOT dataset. The association performance of SparseTrack is substantially contingent on the detector's capacity to accurately identify occluded targets. To be precise, SparseTrack places a higher reliance on the detection quality of partially obscured and low-confidence targets compared to conventional detection-based tracking methods. In addition, SparseTrack faces challenges in other demanding scenarios such as \cite{visdrone, uavdt} where targets exhibit rapid motion and deformation. In such cases, it becomes difficult for the pseudo-depth method to capture the relative depth relationships accurately, which affects the tracking performance of SparseTrack. For videos with low frame rates \cite{bdd100k}, simple Kalman filter motion model struggles to obtain accurate motion cues, which would result in even less accurate pseudo-depth information. In the future work, we will explore the different decomposition methods that can be adapted to various tracking scenarios and aim to make the implementation more elegant.}

\section{Conclusion}\label{conclusion}
We propose a simple tracker for multi-object tracking named as SparseTrack. It leverages the pseudo-depth method to estimate the relative depth relationship between different targets and divides the target set into multiple sparse subsets in order of increasing depth. In order to associate occluded targets distributed across these sparse subsets, we introduce the Depth Cascade Matching (DCM) algorithm, which performs the association between detection subsets and trajectory subsets at the same depth level. Compared to previous tracking methods, SparseTrack offers a different perspective on addressing occlusion: the target set decomposition, which partitions the dense target set into sparse subsets by pseudo-depth information. SparseTrack achieves competitive performance on the MOT17 and MOT20 datasets comparing to state-of-the-art methods via solely utilizing simple IoU distance association, without relying on robust appearance embeddings or enhanced motion prediction. This demonstrates the effectiveness of decomposition based on depth information. It is worth noting that DCM is plug-and-play and can be integrated into any existing tracker, yielding consistent performance improvements. We hope that SparseTrack can provide alternative solutions for multi-object tracking tasks and inspire the development of more powerful and elegant approaches based on the concept of \textbf{sparsity} in the future.
\section*{Acknowledgments}
We thank Yifu Zhang, Bencheng Liao, Jiemin Fang, Yunchi Zhang, and Jinfeng Yao for their insightful discussions and suggestions. This work is in part supported by the National Key Research and Development Program of China under Grant  2022YFB4500602.

\bibliographystyle{IEEEtran}
\bibliography{6_ref}

\begin{thebibliography}{10}
\providecommand{\url}[1]{#1}
\csname url@samestyle\endcsname
\providecommand{\newblock}{\relax}
\providecommand{\bibinfo}[2]{#2}
\providecommand{\BIBentrySTDinterwordspacing}{\spaceskip=0pt\relax}
\providecommand{\BIBentryALTinterwordstretchfactor}{4}
\providecommand{\BIBentryALTinterwordspacing}{\spaceskip=\fontdimen2\font plus
\BIBentryALTinterwordstretchfactor\fontdimen3\font minus
  \fontdimen4\font\relax}
\providecommand{\BIBforeignlanguage}[2]{{%
\expandafter\ifx\csname l@#1\endcsname\relax
\typeout{** WARNING: IEEEtran.bst: No hyphenation pattern has been}%
\typeout{** loaded for the language `#1'. Using the pattern for}%
\typeout{** the default language instead.}%
\else
\language=\csname l@#1\endcsname
\fi
#2}}
\providecommand{\BIBdecl}{\relax}
\BIBdecl

\bibitem{vandenhende2021multi}
S.~Vandenhende, S.~Georgoulis, W.~Van~Gansbeke, M.~Proesmans, D.~Dai, and
  L.~Van~Gool, ``Multi-task learning for dense prediction tasks: A survey,''
  \emph{IEEE Transactions on Pattern Analysis and Machine Intelligence}, 2021.

\bibitem{mot15}
\BIBentryALTinterwordspacing
L.~Leal-Taix\'{e}, A.~Milan, I.~Reid, S.~Roth, and K.~Schindler,
  ``{MOTC}hallenge 2015: {T}owards a benchmark for multi-target tracking,''
  \emph{arXiv:1504.01942 [cs]}, Apr. 2015, arXiv: 1504.01942. [Online].
  Available: \url{http://arxiv.org/abs/1504.01942}
\BIBentrySTDinterwordspacing

\bibitem{mot16}
A.~Milan, L.~Leal-Taix\'{e}, I.~Reid, S.~Roth, and K.~Schindler, ``Mot16: A
  benchmark for multi-object tracking,'' \emph{arXiv preprint
  arXiv:1603.00831}, 2016.

\bibitem{mot20}
P.~Dendorfer, H.~Rezatofighi, A.~Milan, J.~Shi, D.~Cremers, I.~Reid, S.~Roth,
  K.~Schindler, and L.~Leal-Taix\'{e}, ``Mot20: A benchmark for multi object
  tracking in crowded scenes,'' \emph{arXiv preprint arXiv:2003.09003}, 2020.

\bibitem{Bewley2016_sort}
A.~Bewley, Z.~Ge, L.~Ott, F.~Ramos, and B.~Upcroft, ``Simple online and
  realtime tracking,'' in \emph{2016 IEEE International Conference on Image
  Processing (ICIP)}, 2016, pp. 3464--3468.

\bibitem{bytetrack}
Y.~Zhang, P.~Sun, Y.~Jiang, D.~Yu, F.~Weng, Z.~Yuan, P.~Luo, W.~Liu, and
  X.~Wang, ``Bytetrack: Multi-object tracking by associating every detection
  box,'' 2022.

\bibitem{transcenter}
Y.~Xu, Y.~Ban, G.~Delorme, C.~Gan, D.~Rus, and X.~Alameda-Pineda,
  ``Transcenter: Transformers with dense queries for multiple-object
  tracking,'' \emph{arXiv preprint arXiv:2103.15145}, 2021.

\bibitem{transtrack}
P.~Sun, Y.~Jiang, R.~Zhang, E.~Xie, J.~Cao, X.~Hu, T.~Kong, Z.~Yuan, C.~Wang,
  and P.~Luo, ``Transtrack: Multiple-object tracking with transformer,''
  \emph{arXiv preprint arXiv:2012.15460}, 2020.

\bibitem{transmot}
P.~Chu, J.~Wang, Q.~You, H.~Ling, and Z.~Liu, ``Transmot: Spatial-temporal
  graph transformer for multiple object tracking,'' \emph{arXiv preprint
  arXiv:2104.00194}, 2021.

\bibitem{p3aformer}
Z.~Zhao, Z.~Wu, Y.~Zhuang, B.~Li, and J.~Jia, ``Tracking objects as pixel-wise
  distributions,'' 2022.

\bibitem{motr}
F.~Zeng, B.~Dong, Y.~Zhang, T.~Wang, X.~Zhang, and Y.~Wei, ``Motr: End-to-end
  multiple-object tracking with transformer,'' in \emph{European Conference on
  Computer Vision (ECCV)}, 2022.

\bibitem{motrv2}
Y.~Zhang, T.~Wang, and X.~Zhang, ``Motrv2: Bootstrapping end-to-end
  multi-object tracking by pretrained object detectors,'' \emph{Computer Vision
  and Pattern Recognition CVPR}, 2023.

\bibitem{girshick14CVPR}
R.~Girshick, J.~Donahue, T.~Darrell, and J.~Malik, ``Rich feature hierarchies
  for accurate object detection and semantic segmentation,'' in \emph{Computer
  Vision and Pattern Recognition (CVPR)}, 2014.

\bibitem{hota}
J.~Luiten, A.~Osep, P.~Dendorfer, P.~Torr, A.~Geiger, L.~Leal-Taix{\'e}, and
  B.~Leibe, ``Hota: A higher order metric for evaluating multi-object
  tracking,'' \emph{International journal of computer vision}, vol. 129, no.~2,
  pp. 548--578, 2021.

\bibitem{dancetrack}
P.~Sun, J.~Cao, Y.~Jiang, Z.~Yuan, S.~Bai, K.~Kitani, and P.~Luo, ``Dancetrack:
  Multi-object tracking in uniform appearance and diverse motion,'' \emph{arXiv
  preprint arXiv:2111.14690}, 2021.

\bibitem{ioutracker}
\BIBentryALTinterwordspacing
E.~Bochinski, V.~Eiselein, and T.~Sikora, ``High-speed tracking-by-detection
  without using image information,'' in \emph{International Workshop on Traffic
  and Street Surveillance for Safety and Security at IEEE AVSS 2017}, Lecce,
  Italy, Aug. 2017. [Online]. Available:
  \url{http://elvera.nue.tu-berlin.de/files/1517Bochinski2017.pdf}
\BIBentrySTDinterwordspacing

\bibitem{kf}
K.~RE, ``A new approach to linear filtering and prediction problems.'' \emph{J
  Fluids Eng}, vol.~82, no.~1, pp. 35--45, 1960.

\bibitem{deepsort}
N.~Wojke, A.~Bewley, and D.~Paulus, ``Simple online and realtime tracking with
  a deep association metric,'' in \emph{2017 IEEE International Conference on
  Image Processing (ICIP)}.\hskip 1em plus 0.5em minus 0.4em\relax IEEE, 2017,
  pp. 3645--3649.

\bibitem{motdt}
C.~Long, A.~Haizhou, Z.~Zijie, and S.~Chong, ``Real-time multiple people
  tracking with deeply learned candidate selection and person
  re-identification,'' in \emph{ICME}, 2018.

\bibitem{fastreid}
L.~He, X.~Liao, W.~Liu, X.~Liu, P.~Cheng, and T.~Mei, ``Fastreid: A pytorch
  toolbox for general instance re-identification,'' \emph{arXiv preprint
  arXiv:2006.02631}, 2020.

\bibitem{torchreid}
K.~Zhou and T.~Xiang, ``Torchreid: A library for deep learning person
  re-identification in pytorch,'' \emph{arXiv preprint arXiv:1910.10093}, 2019.

\bibitem{zhou2019osnet}
K.~Zhou, Y.~Yang, A.~Cavallaro, and T.~Xiang, ``Omni-scale feature learning for
  person re-identification,'' in \emph{ICCV}, 2019.

\bibitem{zhou2021osnet}
------, ``Learning generalisable omni-scale representations for person
  re-identification,'' 2021.

\bibitem{unitrack}
Z.~Wang, H.~Zhao, Y.-L. Li, S.~Wang, P.~Torr, and L.~Bertinetto, ``Do different
  tracking tasks require different appearance models?'' \emph{Thirty-Fifth
  Conference on Neural Infromation Processing Systems}, 2021.

\bibitem{qdtrack}
J.~Pang, L.~Qiu, X.~Li, H.~Chen, Q.~Li, T.~Darrell, and F.~Yu, ``Quasi-dense
  similarity learning for multiple object tracking,'' in \emph{IEEE/CVF
  Conference on Computer Vision and Pattern Recognition}, June 2021.

\bibitem{UTM}
S.~You, H.~Yao, B.-K. Bao, and C.~Xu, ``Utm: A unified multiple object tracking
  model with identity-aware feature enhancement,'' \emph{Computer Vision and
  Pattern Recognition (CVPR)}, 2023.

\bibitem{semi}
W.~Li, Y.~Xiong, S.~Yang, M.~Xu, Y.~Wang, and W.~Xia, ``Semi-tcl:
  Semi-supervised track contrastive representation learning,'' \emph{arXiv
  preprint arXiv:2107.02396}, 2021.

\bibitem{online_mot}
\BIBentryALTinterwordspacing
Q.~Liu, D.~Chen, Q.~Chu, L.~Yuan, B.~Liu, L.~Zhang, and N.~Yu, ``Online
  multi-object tracking with unsupervised re-identification learning and
  occlusion estimation,'' \emph{Neurocomput.}, vol. 483, no.~C, p. 333–347,
  apr 2022. [Online]. Available:
  \url{https://doi.org/10.1016/j.neucom.2022.01.008}
\BIBentrySTDinterwordspacing

\bibitem{moco}
K.~He, H.~Fan, Y.~Wu, S.~Xie, and R.~Girshick, ``Momentum contrast for
  unsupervised visual representation learning,'' \emph{arXiv preprint
  arXiv:1911.05722}, 2019.

\bibitem{mocov2}
X.~Chen, H.~Fan, R.~Girshick, and K.~He, ``Improved baselines with momentum
  contrastive learning,'' \emph{arXiv preprint arXiv:2003.04297}, 2020.

\bibitem{simclr}
T.~Chen, S.~Kornblith, M.~Norouzi, and G.~Hinton, ``A simple framework for
  contrastive learning of visual representations,'' \emph{arXiv preprint
  arXiv:2002.05709}, 2020.

\bibitem{simclrv2}
T.~Chen, S.~Kornblith, K.~Swersky, M.~Norouzi, and G.~Hinton, ``Big
  self-supervised models are strong semi-supervised learners,'' \emph{arXiv
  preprint arXiv:2006.10029}, 2020.

\bibitem{MAE}
K.~He, X.~Chen, S.~Xie, Y.~Li, P.~Doll{\'a}r, and R.~Girshick, ``Masked
  autoencoders are scalable vision learners,'' \emph{arXiv:2111.06377}, 2021.

\bibitem{jde}
Z.~Wang, L.~Zheng, Y.~Liu, and S.~Wang, ``Towards real-time multi-object
  tracking,'' \emph{The European Conference on Computer Vision (ECCV)}, 2020.

\bibitem{fairmot}
Y.~Zhang, C.~Wang, X.~Wang, W.~Zeng, and W.~Liu, ``Fairmot: On the fairness of
  detection and re-identification in multiple object tracking,''
  \emph{International Journal of Computer Vision}, vol. 129, pp. 3069--3087,
  2021.

\bibitem{CStrack}
C.~Liang, Z.~Zhang, X.~Zhou, B.~Li, S.~Zhu, and W.~Hu, ``Rethinking the
  competition between detection and reid in multiobject tracking,'' \emph{IEEE
  Trans Image Process}, pp. 3182--3196, 2022.

\bibitem{trackformer}
T.~Meinhardt, A.~Kirillov, L.~Leal-Taixe, and C.~Feichtenhofer, ``Trackformer:
  Multi-object tracking with transformers,'' in \emph{The IEEE Conference on
  Computer Vision and Pattern Recognition (CVPR)}, June 2022.

\bibitem{attention}
A.~Vaswani, N.~Shazeer, N.~Parmar, J.~Uszkoreit, L.~Jones, A.~N. Gomez,
  {\L}.~Kaiser, and I.~Polosukhin, ``Attention is all you need,'' in
  \emph{Advances in neural information processing systems}, 2017, pp.
  5998--6008.

\bibitem{vit}
\BIBentryALTinterwordspacing
A.~Dosovitskiy, L.~Beyer, A.~Kolesnikov, D.~Weissenborn, X.~Zhai,
  T.~Unterthiner, M.~Dehghani, M.~Minderer, G.~Heigold, S.~Gelly, J.~Uszkoreit,
  and N.~Houlsby, ``An image is worth 16x16 words: Transformers for image
  recognition at scale,'' in \emph{International Conference on Learning
  Representations}, 2021. [Online]. Available:
  \url{https://openreview.net/forum?id=YicbFdNTTy}
\BIBentrySTDinterwordspacing

\bibitem{detr}
N.~Carion, F.~Massa, G.~Synnaeve, N.~Usunier, A.~Kirillov, and S.~Zagoruyko,
  ``End-to-end object detection with transformers,'' in \emph{European
  conference on computer vision}.\hskip 1em plus 0.5em minus 0.4em\relax
  Springer, 2020, pp. 213--229.

\bibitem{deformable-detr}
\BIBentryALTinterwordspacing
X.~Zhu, W.~Su, L.~Lu, B.~Li, X.~Wang, and J.~Dai, ``Deformable {\{}detr{\}}:
  Deformable transformers for end-to-end object detection,'' in
  \emph{International Conference on Learning Representations}, 2021. [Online].
  Available: \url{https://openreview.net/forum?id=gZ9hCDWe6ke}
\BIBentrySTDinterwordspacing

\bibitem{bdd100k}
F.~Yu, H.~Chen, X.~Wang, W.~Xian, Y.~Chen, F.~Liu, V.~Madhavan, and T.~Darrell,
  ``Bdd100k: A diverse driving dataset for heterogeneous multitask learning,''
  in \emph{Proceedings of the IEEE/CVF conference on computer vision and
  pattern recognition}, 2020, pp. 2636--2645.

\bibitem{dt}
C.~Feichtenhofer, A.~Pinz, and A.~Zisserman, ``Detect to track and track to
  detect,'' in \emph{International Conference on Computer Vision (ICCV)}, 2017.

\bibitem{centertrack}
X.~Zhou, V.~Koltun, and P.~Kr{\"a}henb{\"u}hl, ``Tracking objects as points,''
  \emph{ECCV}, 2020.

\bibitem{siammot}
B.~Shuai, A.~Berneshawi, X.~Li, D.~Modolo, and J.~Tighe, ``Siammot: Siamese
  multi-object tracking,'' in \emph{Proceedings of the IEEE/CVF Conference on
  Computer Vision and Pattern Recognition}, 2021, pp. 12\,372--12\,382.

\bibitem{Tracktor}
P.~Bergmann, T.~Meinhardt, and L.~Leal{-}Taix{\'{e}}, ``Tracking without bells
  and whistles,'' in \emph{The IEEE International Conference on Computer Vision
  (ICCV)}, October 2019.

\bibitem{ctracker}
J.~Peng, C.~Wang, F.~Wan, Y.~Wu, Y.~Wang, Y.~Tai, C.~Wang, J.~Li, F.~Huang, and
  Y.~Fu, ``Chained-tracker: Chaining paired attentive regression results for
  end-to-end joint multiple-object detection and tracking,'' in
  \emph{Proceedings of the European Conference on Computer Vision}, 2020.

\bibitem{cnn}
A.~Krizhevsky, I.~Sutskever, and G.~E. Hinton, ``Imagenet classification with
  deep convolutional neural networks,'' \emph{NIPS}, 2012.

\bibitem{sushi}
O.~Cetintas, G.~Bras'o, and L.~Leal-Taix'e, ``Unifying short and long-term
  tracking with graph hierarchies,'' in \emph{Proceedings of the IEEE/CVF
  Conference on Computer Vision and Pattern Recognition (CVPR)}, June 2023, pp.
  22\,877--22\,887.

\bibitem{graph-mot}
G.~Braso and L.~Leal-Taixe, ``Learning a neural solver for multiple object
  tracking,'' \emph{CVPR}, 2020.

\bibitem{Learnable_graph_matching}
J.~He, Z.~Huang, N.~Wang, and Z.~Zhang, ``Learnable graph matching:
  Incorporating graph partitioning with deep feature learning for multiple
  object tracking,'' 2021, p. 5299–5309.

\bibitem{graph-mot1}
J.~Li, X.~Gao, and T.~Jiang, ``Graph networks for multiple object tracking,''
  \emph{In Proceedings of the IEEE/CVF Winter Conference on Applications of
  Computer Vision (WACV)}, March 2020.

\bibitem{GSDT}
Y.~Wang, K.~Kitani, and X.~Weng, ``Joint object detection and multi-object
  tracking with graph neural networks,'' \emph{arXiv preprint
  arXiv:2006.13164}, 2020.

\bibitem{motiontrack}
Z.~Qin, S.~Zhou, L.~Wang, J.~Duan, G.~Hua, and W.~Tang, ``Motiontrack: Learning
  robust short-term and long-term motions for multi-object tracking,''
  \emph{Computer Vision and Pattern Recognition (CVPR)}, 2023.

\bibitem{dp_mot}
K.~G. Quach, H.~Le, P.~Nguyen, C.~N. Duong, T.~D. Bui, and K.~Luu, ``Depth
  perspective-aware multiple object tracking,'' 2023.

\bibitem{BoT-SORT}
N.~Aharon, R.~Orfaig, and B.-Z. Bobrovsky, ``Bot-sort: Robust associations
  multi-pedestrian tracking,'' \emph{arXiv preprint arXiv:2206.14651}, 2022.

\bibitem{aplift}
A.~Hornakova, T.~Kaiser, P.~Swoboda, M.~Rolinek, B.~Rosenhahn, and R.~Henschel,
  ``Making higher order mot scalable: An efficient approximate solver for
  lifted disjoint paths,'' in \emph{Proceedings of the IEEE/CVF International
  Conference on Computer Vision (ICCV)}, October 2021, pp. 6330--6340.

\bibitem{9857481}
Y.~Liu, X.~Zhang, B.~Zhang, X.~Zhang, S.~Wang, and J.~Xu, ``Multi-camera
  vehicle tracking based on occlusion-aware and inter-vehicle information,'' in
  \emph{2022 IEEE/CVF Conference on Computer Vision and Pattern Recognition
  Workshops (CVPRW)}, 2022.

\bibitem{LMGP}
D.~H. Nguyen, R.~Henschel, B.~Rosenhahn, D.~Sonntag, and P.~Swoboda, ``Lmgp:
  Lifted multicut meets geometry projections for multi-camera multi-object
  tracking,'' in \emph{2022 IEEE/CVF Conference on Computer Vision and Pattern
  Recognition (CVPR)}, 2022.

\bibitem{nerf}
B.~Mildenhall, P.~P. Srinivasan, M.~Tancik, J.~T. Barron, R.~Ramamoorthi, and
  R.~Ng, ``Nerf: Representing scenes as neural radiance fields for view
  synthesis,'' in \emph{ECCV}, 2020.

\bibitem{trackrcnn}
P.~Voigtlaender, M.~Krause, A.~Osep, J.~Luiten, B.~B.~G. Sekar, A.~Geiger, and
  B.~Leibe, ``{MOTS}: Multi-object tracking and segmentation,'' in \emph{CVPR},
  2019.

\bibitem{FANTrack}
E.~Baser, V.~Balasubramanian, P.~Bhattacharyya, and K.~Czarnecki, ``Fantrack:
  3d multi-object tracking with feature association network,'' in \emph{IEEE
  Intelligent Vehicles Symposium (IV 19)}, 2019.

\bibitem{centerpoint}
T.~Yin, X.~Zhou, and P.~Kr{\"a}henb{\"u}hl, ``Center-based 3d object detection
  and tracking,'' \emph{CVPR}, 2021.

\bibitem{yolox}
Z.~Ge, S.~Liu, F.~Wang, Z.~Li, and J.~Sun, ``Yolox: Exceeding yolo series in
  2021,'' \emph{arXiv preprint arXiv:2107.08430}, 2021.

\bibitem{strongsort}
Y.~Du, Y.~Song, B.~Yang, and Y.~Zhao, ``Strongsort: Make deepsort great
  again,'' \emph{arXiv preprint arXiv:2202.13514}, 2022.

\bibitem{TraDeS}
J.~Wu, J.~Cao, L.~Song, Y.~Wang, M.~Yang, and J.~Yuan, ``Track to detect and
  segment: An online multi-object tracker,'' in \emph{Proceedings of the
  IEEE/CVF Conference on Computer Vision and Pattern Recognition}, 2021, pp.
  12\,352--12\,361.

\bibitem{CLEAR-metrics}
K.~Bernardin and R.~Stiefelhagen, ``Evaluating multiple object tracking
  performance: the clear mot metrics,'' \emph{EURASIP Journal on Image and
  Video Processing}, vol. 2008, pp. 1--10, 2008.

\bibitem{idf1}
E.~Ristani, F.~Solera, R.~Zou, R.~Cucchiara, and C.~Tomasi, ``Performance
  measures and a data set for multi-target, multi-camera tracking,'' in
  \emph{ECCV}.\hskip 1em plus 0.5em minus 0.4em\relax Springer, 2016, pp.
  17--35.

\bibitem{tubetk}
B.~Pang, Y.~Li, Y.~Zhang, M.~Li, and C.~Lu, ``Tubetk: Adopting tubes to track
  multi-object in a one-step training model,'' in \emph{Proceedings of the
  IEEE/CVF Conference on Computer Vision and Pattern Recognition}, 2020, pp.
  6308--6318.

\bibitem{mat}
S.~Han, P.~Huang, H.~Wang, E.~Yu, D.~Liu, X.~Pan, and J.~Zhao, ``Mat:
  Motion-aware multi-object tracking,'' \emph{arXiv preprint arXiv:2009.04794},
  2020.

\bibitem{PermaTrackPr}
P.~Tokmakov, J.~Li, W.~Burgard, and A.~Gaidon, ``Learning to track with object
  permanence,'' \emph{arXiv preprint arXiv:2103.14258}, 2021.

\bibitem{oc-sort}
J.~Cao, X.~Weng, R.~Khirodkar, J.~Pang, and K.~Kitani, ``Observation-centric
  sort: Rethinking sort for robust multi-object tracking,'' \emph{arXiv
  preprint arXiv:2203.14360}, 2022.

\bibitem{DAN}
S.~Sun, N.~Akhtar, H.~Song, A.~S. Mian, and M.~Shah, ``Deep affinity network
  for multiple object tracking,'' \emph{IEEE transactions on pattern analysis
  and machine intelligence}, 2019.

\bibitem{SOTMOT}
L.~Zheng, M.~Tang, Y.~Chen, G.~Zhu, J.~Wang, and H.~Lu, ``Improving multiple
  object tracking with single object tracking,'' in \emph{Proceedings of the
  IEEE/CVF Conference on Computer Vision and Pattern Recognition}, 2021, pp.
  2453--2462.

\bibitem{remot}
F.~Yang, X.~Chang, S.~Sakti, Y.~Wu, and S.~Nakamura, ``Remot: A model-agnostic
  refinement for multiple object tracking,'' \emph{Image and Vision Computing},
  vol. 106, p. 104091, 2021.

\bibitem{relationtrack}
E.~Yu, Z.~Li, S.~Han, and H.~Wang, ``Relationtrack: Relation-aware multiple
  object tracking with decoupled representation,'' \emph{arXiv preprint
  arXiv:2105.04322}, 2021.

\bibitem{FUFET}
C.~Shan, C.~Wei, B.~Deng, J.~Huang, X.-S. Hua, X.~Cheng, and K.~Liang,
  ``Tracklets predicting based adaptive graph tracking,'' \emph{arXiv preprint
  arXiv:2010.09015}, 2020.

\bibitem{CorrTrack}
Q.~Wang, Y.~Zheng, P.~Pan, and Y.~Xu, ``Multiple object tracking with
  correlation learning,'' in \emph{Proceedings of the IEEE/CVF Conference on
  Computer Vision and Pattern Recognition}, 2021, pp. 3876--3886.

\bibitem{MLT}
Y.~Zhang, H.~Sheng, Y.~Wu, S.~Wang, W.~Ke, and Z.~Xiong, ``Multiplex labeling
  graph for near-online tracking in crowded scenes,'' \emph{IEEE Internet of
  Things Journal}, vol.~7, no.~9, pp. 7892--7902, 2020.

\bibitem{FCG}
A.~Girbau, F.~Marqués, and S.~Satoh, ``Multiple object tracking from
  appearance by hierarchically clustering tracklets,'' \emph{BMVC}, 2022.

\bibitem{GTR}
X.~Zhou, T.~Yin, V.~Koltun, and P.~Kr{\"a}henb{\"u}hl, ``Global tracking
  transformers,'' in \emph{CVPR}, 2022.

\bibitem{visdrone}
P.~Zhu, L.~Wen, D.~Du, X.~Bian, H.~Fan, Q.~Hu, and H.~Ling, ``Detection and
  tracking meet drones challenge,'' \emph{IEEE Transactions on Pattern Analysis
  and Machine Intelligence}, pp. 1--1, 2021.

\bibitem{uavdt}
D.~Du, Y.~Qi, H.~Yu, Y.~Yang, K.~Duan, G.~Li, W.~Zhang, Q.~Huang, and Q.~Tian,
  ``The unmanned aerial vehicle benchmark: Object detection and tracking,''
  \emph{ECCV}, pp. 370--386, 2018.

\end{thebibliography}

\newpage

\end{document}